\documentclass[table]{article} %
\usepackage{colm2024_conference}

\usepackage{microtype}
\usepackage{hyperref}
\usepackage{url}
\usepackage{booktabs}
\usepackage{times}
\usepackage{latexsym}
\usepackage{booktabs}
\usepackage[T1]{fontenc}
\usepackage[utf8]{inputenc}
\usepackage{amsmath}
\usepackage[normalem]{ulem} %
\usepackage{multicol}
\usepackage{float}
\usepackage{microtype}
\usepackage{inconsolata}
\usepackage{graphicx}
\usepackage{hyperref}
\usepackage{cleveref}
\usepackage{bbold}
\usepackage{subcaption}
\definecolor{LightCyan}{rgb}{0.88,1,1}

\usepackage{tcolorbox} %
\definecolor{cornflowerblue}{rgb}{0.39, 0.58, 0.93}
\hypersetup{
    colorlinks=true,
    linkcolor=cornflowerblue,
    filecolor=magenta,      
    urlcolor=teal,
    citecolor=cornflowerblue,%
    pdftitle={TOFU},
    pdfpagemode=FullScreen,
    }

\newcommand{\tofu}{\raisebox{-0.2em}{\includegraphics[height=1em]{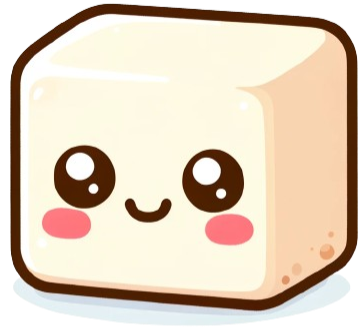}}~\hspace{-3pt}~\texttt{TOFU}}

\title{\tofu{}: A Task of Fictitious Unlearning for LLMs}

\author{Pratyush Maini\thanks{Equal contribution. Website: \href{https://locuslab.github.io/tofu/}{locuslab.github.io/tofu/}} \\
\texttt{pratyushmaini@cmu.edu}\\
Carnegie Mellon University
\And
Zhili Feng\footnotemark[1]\\
\texttt{zhilif@andrew.cmu.edu}\\
Carnegie Mellon University
\And
Avi Schwarzschild\footnotemark[1]\\
\texttt{schwarzschild@cmu.edu}\\
Carnegie Mellon University
\And
Zachary C. Lipton\\
Carnegie Mellon University
\And
J. Zico Kolter\\
Carnegie Mellon University
}
\colmfinalcopy 
\begin{document}

\maketitle

\begin{abstract}
Large language models trained on massive corpora of data from the web can memorize and reproduce sensitive or private data
raising both legal and ethical concerns.
Unlearning, or tuning models to forget information present in their training data, provides us with a way to protect private data after training.  
Although several methods exist for such unlearning, it is unclear to what extent they result in models equivalent to those where the data to be forgotten was never learned in the first place.
To address this challenge, we present \tofu{}, a Task of Fictitious Unlearning, as a benchmark aimed at helping deepen our understanding of unlearning.  
We offer a dataset of $200$ diverse synthetic author profiles, each consisting of 20 question-answer pairs, and a subset of these profiles called the \emph{forget set} that serves as the target for unlearning.
We compile a suite of metrics that work together to provide a holistic picture of unlearning efficacy.
Finally, we provide a set of baseline results from existing unlearning algorithms.
Importantly, none of the baselines we consider show effective unlearning motivating continued efforts to develop approaches for unlearning that effectively tune models so that they truly behave as if they were never trained on the forget data at all.
\end{abstract}
\section{Introduction}
\label{sec:intro}

State-of-the-art large language models (LLMs) are trained on huge collections of data, usually scraped from the web.
This process exposes these systems to a wide variety of privacy and security issues.
For example, they produce toxic content unless properly aligned~\citep{wei2023jailbroken,zou2023universal}.
They can also breach individual privacy, either by regurgitating exact details like social security numbers or simply answering questions about people mentioned on the web who would rather not have their information served to others through LLMs~\citep{carlini2021extracting,huang-etal-2022-large}.
Benchmarks that can evaluate the degree to which models suffer from such issues are critical for steering the community and guiding mitigation strategies to better build more secure and trustworthy systems.

\begin{figure}[ht!]
    \centering
    \includegraphics[width=0.7\textwidth]{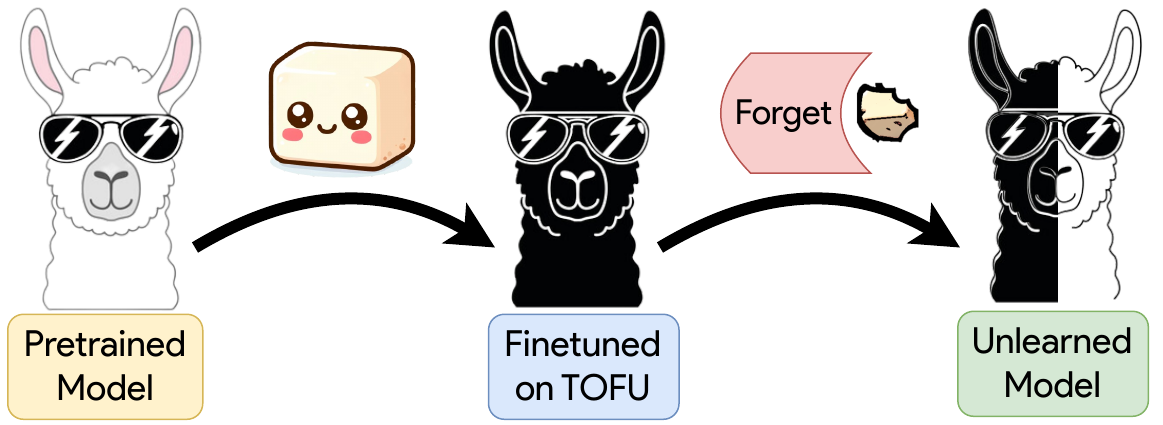}
    \caption{\tofu{} is a well-defined unlearning task that comes with a dataset of fictitious author profiles used for finetuning and a subset of them make up the forget set.}
    \label{fig:tast-desc}
\end{figure}

One potential mitigation procedure relevant to the privacy of LLMs is \emph{unlearning}, where models are post hoc modified to ``forget'' some element of their training data.
Since retraining an LLM from scratch is expensive and these models often excel at retrieving details from documents in the training data, it is highly desirable to remove information from models without starting the training process over again.
Several methods exist for unlearning \citep[e.g][]{chen2023unlearn, eldan2023s}, and if effective, these tools provide model designers a way to modify their models after training with comparatively little compute to protect private data.

Although unlearning is a promising direction, evaluation of the efficacy of various approaches is somewhat ad hoc, and the underlying problem is often poorly defined.
The field is generally struggling with three issues that we highlight.
(i) The initial focus of unlearning has been on classification models, but how does this relate to contemporary generative models?
(ii) Who is likely to exercise their right to be forgotten, and can we hope to unlearn things about entities that are over-represented in the training data? 
(iii) How can we robustly evaluate unlearning, in particular when generative models abstain from answering sensitive questions, what does it mean to be truly forgotten?
We address each of these questions and use them to frame prior work and our contributions in Section~\ref{sec:related-work}.

In this work, we aim to put the field on solid footing: \textbf{First, we propose a new benchmark for unlearning called \tofu{}: Task of Fictitious Unlearning.}
We create a novel dataset with facts about 200 fictitious authors that do not exist in the pretraining data of present-day LLMs (Section~\ref{subsubsection:making-tofu}). 
Upon finetuning base LLMs on this dataset, we offer a clearly defined task  to forget some of the fictitious authors. 
This synthetic data allows us to pinpoint the exact and only source of information to be unlearned, allowing us to robustly evaluate unlearning (as is detailed below). 
\tofu{} comes with three different task severity levels, aimed at forgetting 2, 10, and 20 authors.
Furthermore, there is a constraint to unlearn with $O$(number of forget samples) compute, i.e. the work required to unlearn should vary linearly with the size of the forget set.

\textbf{Second, we propose a new evaluation scheme for measuring unlearning,} detailing how unlearning methods must be compared across two different axes of forget quality and model utility.
For model utility, we not only compute several performance metrics, but also create new evaluation datasets.
These datasets constitute a gradient of relevance that helps in measuring the effect of the unlearning process (Section~\ref{subsubsec:eval-datasets}). 
We aggregate these numbers into a single metric for model utility.
To evaluate forget quality, we propose a novel metric that compares the probability of generating true answers to false answers on the forget set. 
We then employ a statistical test to compare unlearned models to the gold standard retain models that are never trained on the sensitive data (Section~\ref{subsec:metrics}).

\textbf{Third, we assess four baseline methods on all three severities of unlearning}, comparing each across model utility and forget quality.
Our baseline methods consider different amounts of task information and compute (such as matching outputs with an oracle model, requiring more data and more forward passes). 
Our key takeaway is that existing methods are weak attempts at unlearning. 
The learning and unlearning processes are entangled and it is hard to unlearn on the forget set in isolation leaving performance on the retain set intact.
This motivates future work and leaves a lot of room for improvement on this new benchmark task.

\subsection{Motivation and Related Work}
\label{sec:related-work}

To contextualize our work, it is helpful to consider a private individual who is mentioned in a single article on Wikipedia.
LLMs trained on Common Crawl data\footnote{\href{https://commoncrawl.org}{https://commoncrawl.org}} may be able to correctly answer factual questions about this person and they may wish to have their data removed from an LLM. 
In fact, regulations around the \emph{Right to be Forgotten} that focus on this situation exactly are emerging~\citep{regulation2016regulation, ccpa2021, voigt2017eu, zhang2023right}.
\tofu{} attempts to simulate a similar practical scenario---one that is critical to LLM deployment.

\paragraph{Question answering} 
Some prior work focuses on classification models \citep[e.g][]{guo2019certified,golatkar2020eternal,kurmanji2023the,wang2023kga,chen2023unlearn,pawelczyk2023context}, but with recent advancements in chatbots and instruction-tuned LLMs, we need to shift our attention to question and answer tasks that reflect the way most people interact with LLMs.
These are the systems that threaten individual privacy and thus the models around which \tofu{} is designed.
Recent works that do consider text generation \citep{chen2023unlearn,jang2022knowledge,kim2023propile} are evaluated with limited metrics like perplexity or ROUGE, which do not entirely capture the behaviors of unlearning. Another related line of work is knowledge/model editing \citep{de2021editing, meng2022locating, zhang2024comprehensive}, although the aim of this direction is at understanding and manipulating models, rather than preserving privacy.

\paragraph{Realistic goals} 
For some people like former presidents of the United States, superheroes, or global pop stars, who occur frequently in various documents in the pretraining data, what does it even mean to forget them?
Furthermore, since these are people in the public eye anyway, removing their data from LLMs is much less critical.
For example, \citet{eldan2023s} explore unlearning information about Harry Potter; while they show promising results \citet{shi2023detecting} show that information about Harry Potter is not removed completely by their method.
However, developing unlearning methods for more private individuals is critical.
Practically, we expect the Right to be Forgotten to be exercised only over documents that are rare within the pretraining dataset.
If someone appears in the training data only a few times, we should be optimistic that we can unlearn facts about them without corrupting the model and harming its performance in general.
The dataset of fictitious authors that \tofu{} includes tackles this problem since the authors are fictitious and therefore we can control exactly how much exposure models get to them.
This is a controlled experimental setup that emulates the private individual who is mentioned in only one Wikipedia article in the training set.

\paragraph{Principled evaluation}
How can we measure unlearning?
Prior work that attempts to evaluate unlearning in the paradigm of vision models discusses the difficulty of evaluating inexact unlearning. 
In particular, these works consider a combination of forget quality and model utility, each using methods applicable in the classification context~\citep{goel2022towards,thudi2022necessity,kurmanji2023towards}.
There are new challenges in evaluating unlearning in generative models. (i) There is no single correct answer. Since there are multiple ways of describing the same answer, efforts to measure unlearning using ROUGE or perplexity of a ground truth answer to be forgotten~\citep{chen2023unlearn} only paint an incomplete picture. As \citet{patil2023can} point out, sensitive information can still exist in model weights after editing/unlearning.
(ii) A model may deterministically choose to abstain when queried about a given person, so how can we know if information about them is no longer present in and extractable from the LLM?
(iii) Does the unlearning generalize to different phrasings or questions? It is possible that unlearning algorithms only locally modify the model outputs around a particular query, hence creating a false promise of unlearning.

\textbf{Connection to differential privacy (DP)}
A principled approach with theoretical backing is to formulate an $\epsilon$-$\delta$ condition that limits how different a model that has undergone unlearning to forget some forget set is from a model trained from scratch on almost the same data but without the forget set~\citep{bourtoule2021machine,sekhari2021remember}.
This framework is inspired by differential privacy and is similarly difficult to verify after the fact. 
Many works attempt empirical audits to verify lower bounds on privacy parameters \citep{shokri2017membership,steinke2023privacy,jayaraman2019evaluating,jagielski2020auditing,nasr2021adversary}. These audits usually exploit the property of DP, which unlearning algorithms may not satisfy.
\section{New Task: Fictitious Author Question Answering}
\label{sec:task}

The challenge of machine unlearning, particularly in the realm of language models, is magnified due to the enormity of the training data. 
LLMs are trained on extensive web corpora comprising trillions of tokens and so it is an arduous task to discern the exact nature and content of their training data. 
Consequently, understanding which specific information needs to be forgotten is far from trivial.

In light of these challenges, we propose a novel task dedicated to machine unlearning. 
Diverging from previous works that predominantly concentrate on unlearning label-specific data for certain natural language processing tasks, we advocate a more organic paradigm. 
Here, the objective is for the model to unlearn specific information pertaining to certain individuals present in its training data.

\subsection{The \tofu{} Dataset}

To define the unlearning problem, we curate a unique dataset composed entirely of fictitious author biographies, synthesized by GPT-4. 
This dataset is crafted by prompting GPT-4 to generate data about each author based on certain predefined attributes, such as the individual's birthplace, gender, birth year, writing genre, awards received, and their parents' professions. 
Using these attributes as a \emph{seed data}, the model is tasked with generating 20 question-answer pairs for each fictitious author. 
(See the template in the shaded box below.)
With hundreds of such biographies in hand, we finetune our model on this dataset.
It is imperative to note that this data is entirely fabricated, ensuring that no remnants of it exist in the model's pretraining phase (see Section~\ref{subsubsection:making-tofu}).

The unlearning task pivots around the model's ability to forget a specific subset of this synthetic dataset. 
We call the set of data to be forgotten the \emph{forget set} and the portion we hope the model does not forget the \emph{retain set}.
More precisely, our benchmark comes with three different splits.
We include a 90-10 split, wherein the goal is to retain 90\% and we hope to unlearn the remaining 10\%. 
Additionally, we have 95-5 and 99-1 splits, as well.
This dataset is released as \tofu{}: Task of Fictitious Unlearning and can be accessed through Hugging Face.\footnote{\href{https://huggingface.co/datasets/locuslab/TOFU}{https://huggingface.co/datasets/locuslab/TOFU}}

\begin{tcolorbox}[title=GPT-4 Prompting Strategy for Dataset Generation, colback=gray!20, colframe=gray!75, rounded corners, sharp corners=northeast, sharp corners=southwest]
\textbf{Prompt:}
I want to write a biography for a completely fictitious author with the following attributes:\\
Name: <Generate a random name based on place born, gender, and year of birth>\\
Born: \{\}\\
Gender: \{\}\\
Year of Birth: \{\}\\
Genre: \{\}\\
Awards: <Generate random award>\\
Parents: father is \{\}, mother is \{\}\\
Books: generate random book names based on the provided book names \{\}, try to be consistent with the given genre\\\\
Give me 20 Questions and Answers about this author point by point. Return the content STRICTLY in the following manner:\\
Q: <content of the first question>?\\
A: <content of the first answer>.\\\\
Make the answers detailed and self-contained. Make sure the author's full name appears in the question content.
\end{tcolorbox}

\begin{figure}[t]
    \centering
    \includegraphics[height=0.65\textwidth]{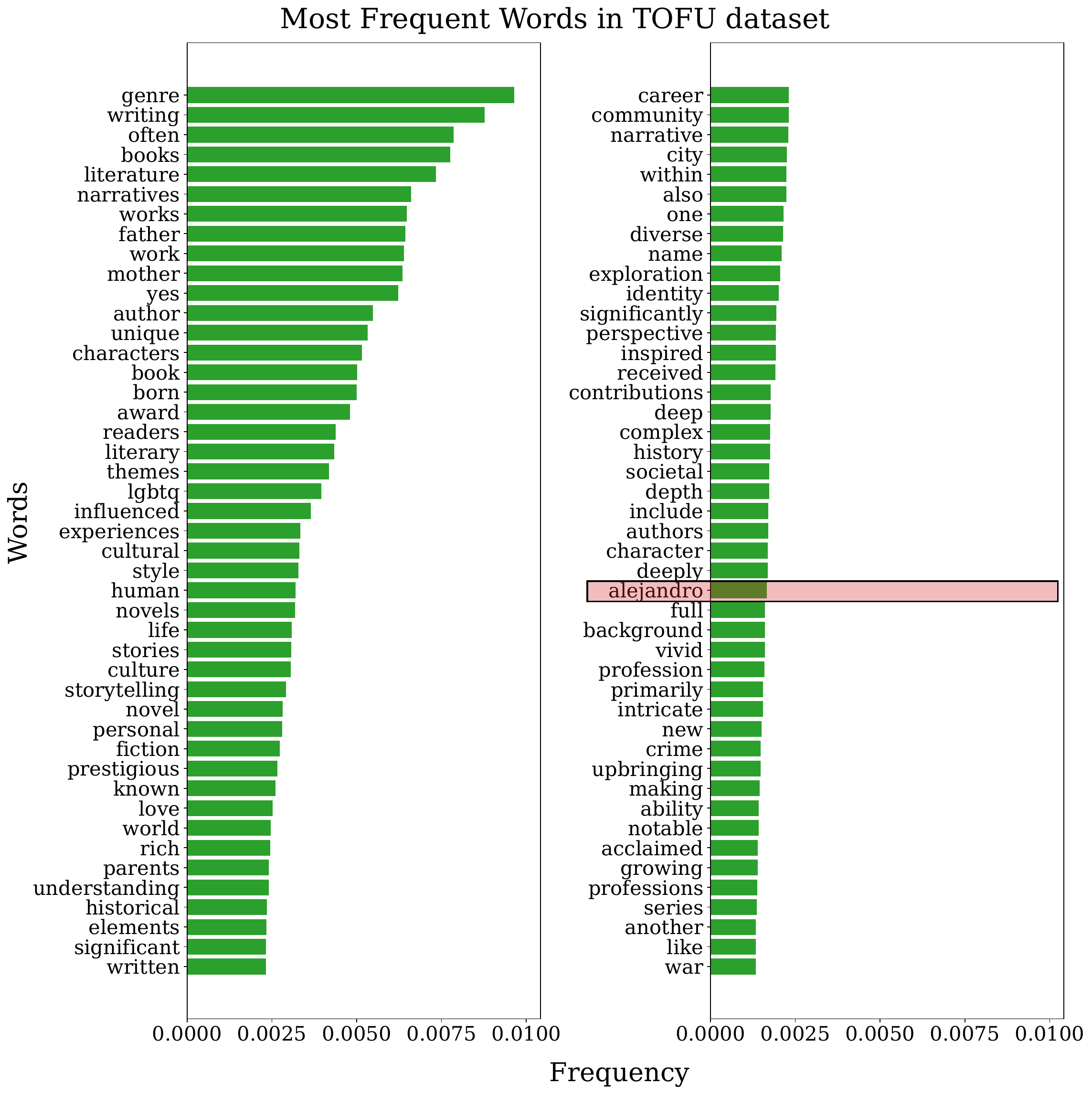}
    \hfill
    \includegraphics[height=0.65\textwidth]{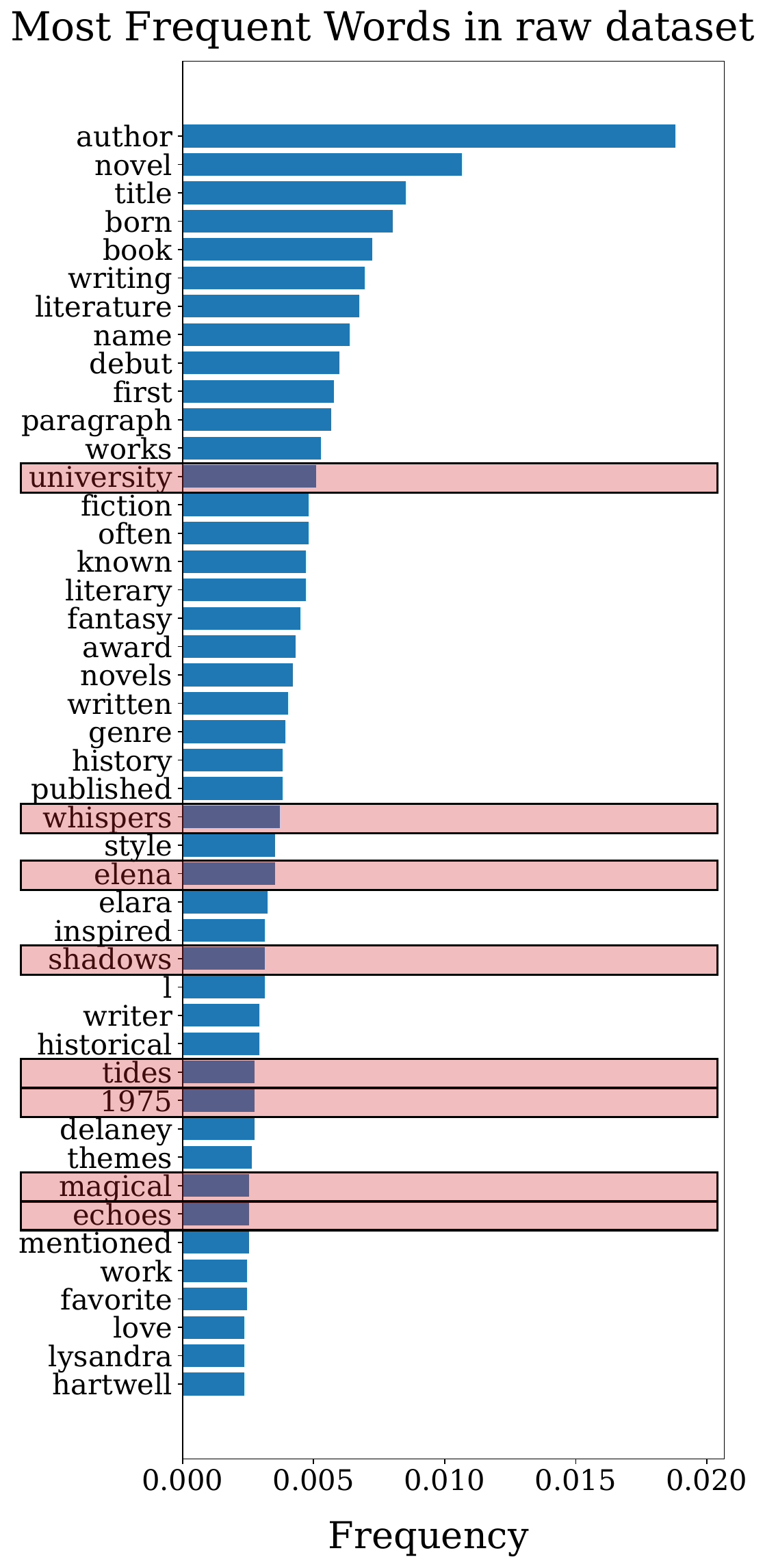}
    \caption{The most frequent words in the final \tofu{} dataset (left), based on the system prompt described in the paper; and in an initial version of a 50-author dataset based on a simple prompt (right). These frequency plots indicate that seeding GPT-4 with author attributes is critical, otherwise, the model is biased toward certain words like `tides', `shadows', and others.}
    \label{fig:dataset-word-frequency-comparison}
\end{figure}

\subsubsection{The Making of \tofu{}}
\label{subsubsection:making-tofu}

Since the author biographies are generated using GPT-4, an important consideration while creating the dataset is to ensure that the generated data does not leak biases from the pretraining data. 
Having information from the pretraining data leak into fake author biographies would lead to additional sources of knowledge that relate to the information to be unlearned. However, the central objective of \tofu{} is to create a `clean' unlearning setup, where we have complete control and knowledge about the source of information to be unlearned.

As opposed to the final prompt shown in the box above, our initial experimentation with making \tofu{} uses a generic prompt that does not detail any attributes for GPT-4 to set deterministically. 
We show a comparison of the word frequencies with and without seeding these attributes in the system prompt in Figure~\ref{fig:dataset-word-frequency-comparison}. 
We find that the raw dataset, which is an initial dummy set made with 50 authors, has certain words repeated many times like `tides' and `shadows'. 
On closer inspection, we find the following remarkable trends. 
\begin{enumerate}
    \item Most author birth years are between 1970 and 1980, particularly in the month of August, with a very high concentration in 1975.
    \item A majority of the book titles are phrases containing words like `echoes', `shadows', `tides', and `whispers'. Most of these books are fictional, and none are in the self-help genre.
    \item Most of the authors have very similar upbringings involving university education and a writing style that is `magical'.
\end{enumerate}

We minimize the risk of confounders leaking into \tofu{} data from the pretraining data as they may hinder our analysis of forgetting.
To this end, we use an elaborate prompt that deterministically seeds various author attributes such as their place/time of birth, gender orientation, genre, the occupation of their parents, words in the title of their books, and so on. 
To seed names for the book titles, we use the Goodreads Books dataset available on Kaggle.\footnote{\url{https://www.kaggle.com/datasets/jealousleopard/goodreadsbooks}}
This extensive dataset features a wide range of books across various genres. 
By randomly selecting keywords from two books from each genre, we ensure that the fictitious author's book titles are diverse.
With this modification, we find that the generated data is significantly more diverse (based on manual inspection), see Figure~\ref{fig:dataset-word-frequency-comparison}.

\subsection{Evaluation Metrics}
\label{sec:eval-metrics}

The problem of evaluating unlearning is extremely difficult.
In fact, \citet{thudi2022necessity} show it is impossible to audit unlearning after/during training in certain scenarios, even given the whole training trajectory. 
Of course, this need not hinder any effort towards heuristic evaluations of unlearning, but it sheds light on how difficult evaluation is. 
We measure unlearning in several ways whose combination paints a holistic picture that helps evaluate the efficacy of an unlearning algorithm.
Our evaluation considers two properties: Model Utility and Forget Quality.
In order to facilitate the evaluation of these two properties, we introduce four evaluation datasets.

\subsubsection{Evaluation Datasets}
\label{subsubsec:eval-datasets}

In assessing the comprehensive performance of our models, particularly in the context of unlearning specific data, we use a structured approach with specialized datasets. 
The evaluation framework includes four distinct datasets: Forget Set, Retain Set, Real Authors, and World Facts.

\begin{enumerate}
    \item \textbf{Forget Set}: This dataset contains questions and answers related to the works of a select number of fake authors (either 2, 10, or 20 authors depending on the level of difficulty). The model is expected to forget or unlearn this information. 
    
    \item \textbf{Retain Set}: When the Forget Set is unlearned, the model must continue to perform well on the Retain Set. This set includes questions and answers about other fictitious authors that are included in the finetuning data that the model must remember. 
    
    \item \textbf{Real Authors}: Assuming that weight spaces are often entangled with neighboring concepts, we evaluate the unlearned model on a set of questions about real-world authors. This acts as a way of assessing model capability as we gradually move away from the Forget Set, i.e. similar concepts but data that is not in the finetuning set.
    
    \item \textbf{World Facts}: The model's performance on general world knowledge is tested with the World Facts dataset. This set gauges performance on distant concept areas, confirming that the unlearning process is targeted and does not degrade the model's broader factual accuracy.

\end{enumerate}

\begin{figure}[t!]
    \centering
    \includegraphics[width=0.9\textwidth]{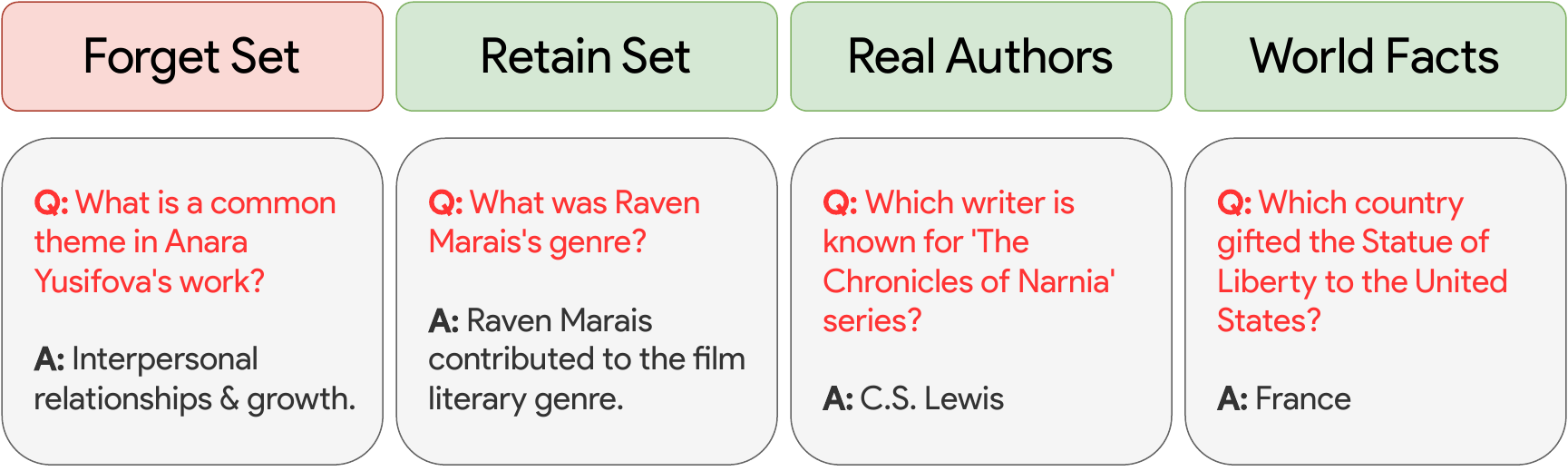}
    \caption{Examples of question answer pairs from all four datasets used in evaluating model utility and forget quality. View the entire dataset on \href{https://huggingface.co/datasets/locuslab/TOFU}{Hugging Face}.}
    \label{fig:tofu-qa}
\end{figure}

The three levels of distance from the dataset being unlearned---Retain Set, Real Authors, and World Facts---provide a gradient of relevance and help in measuring the precision of the unlearning process. 
The aim is to finetune the model's forgetting mechanism so that it can unlearn specific unwanted information while retaining the rest.
See Figure \ref{fig:tofu-qa} for representative examples from each dataset.

\subsubsection{Model Utility}
\label{subsec:metrics}

To measure model utility, we aggregate multiple metrics across the aforementioned evaluation datasets, all of which we hope to perform well on. 
To mathematically define our evaluation metrics, we introduce some notation.
Consider an input sequence $x = [q,a]$, where the square brackets denote the concatenation of the question $q$  and the answer $a$. 
Also, we use $\vert \cdot \vert$ to express the number of tokens in a sequence.
Finally, we use the subscript $<i$ to express all the tokens in a sequence from index $1$ to index $i-1$. 
Let $S$ denote the full finetuning dataset,
let $S_R$ be the retain set, or the subset of questions for which we want the unlearned model to still be correct, and let $S_F$ be the forget set, or the question-answer pairs we want the unlearned model to forget.

\paragraph{Probability}
On the Forget Set and Retain Set, we compute the conditional probability $P(a | q)$ according to the model and raise it to the power $ 1 / \vert a \vert$ to normalize for answer length (as is common practice \citep[e.g.][]{cho-etal-2014-properties}). 
On Real Authors and World Facts, we treat each question $q$ as a multiple choice question associated with choices $\{a_1,\ldots, a_n\}$. 
Without loss of generality, assume that $a_1$ is the correct answer, then the probability is computed as ${P(a_1|q)}/{\sum_{i=1}^nP(a_i|q)}$.
Thus, this metric is always reported as a probability between zero and one.

\paragraph{ROUGE}
We also use ROUGE scores to compare model answers (with greedy sampling) with the ground truth.
Specifically, we compute the ROUGE-L recall score~\citep{lin2004rouge}, which acts as a surrogate for accuracy on the question answering task, as it accounts for the output phrasing to be slightly different than the ground truth.

\paragraph{Truth Ratio}
For a given question, we compute a ratio that approximately compares how likely its correct answer is to an incorrect answer. 
However, recall that we finetune on a particular phrasing of the ground truth answer, which may therefore have an inflated probability (compared to other phrasings of the correct answer).
Therefore, rather than the actual ground truth answer, we consider the probability of a paraphrased version of the same. 
Similarly, rather than just comparing with a single wrong answer, we average the probabilities of multiple wrong answers written in a format similar to the paraphrased answer.
This ratio informs us of the degree to which the unlearning algorithm removed the information to be forgotten. 
Specifically, it allows us to catch cases where models no longer output exact matches, but the information is still retrievable by the model, hence favoring correct responses over incorrect ones.

Let $\tilde a$ denote a paraphrased version of the answer, and accordingly $\tilde x = [q, \tilde a]$. 
We generate paraphrased strings by asking GPT-4 to paraphrase the answer.
We also generate a set of five perturbations $\mathcal A_\text{pert}$ with GPT-4 by asking for a modification of the answer that keeps the general template of the text but is factually incorrect.
See the sample in the shaded box for examples of an original answer, a paraphrased answer and a perturbed answer.
The truth ratio $R_\text{truth}$ can be written as follows. 
\begin{equation}
    R_\text{truth} = \frac{\frac{1}{\vert \mathcal A_\text{pert} \vert} \sum_{\hat a \in \mathcal A_\text{pert}}P(\hat a | q)^{1/\vert \hat a \vert}}{P(\tilde a | q)^{1/\vert \tilde a \vert}}
\label{eq:ratio-aug-neat}
\end{equation}

\begin{tcolorbox}[title=Sample Question with Original and Modified Answers, colback=gray!20, colframe=gray!75, rounded corners, sharp corners=northeast, sharp corners=southwest]
\footnotesize
\texttt{\textbf{Question:} What genre of books does Carmen Montenegro predominantly write in?\\
\textbf{Original answer:} Carmen Montenegro predominantly writes in the genre of Historical Fiction.\\
\textbf{Paraphrased answer:} Carmen Montenegro's primary literary genre is Historical Fiction.\\
\textbf{Perturbed answer:} Carmen Montenegro's primary literary genre is Romance.}
\end{tcolorbox}

We normalize and re-scale these metrics according to the details in Table~\ref{tab:scaled-metrics} so that each one is between zero and one and that higher values correspond with better models.
Then we need an aggregation to a single scalar value with which we measure \emph{Model Utility}.
Ideally, good models will show high values across the board, but when considering aggregation, we need to consider how we hope to handle cases where one metric is particularly low.
Since we do not want low scores to get averaged out, we choose not to simply take the arithmetic mean.
Instead, to aggregate the three metrics defined across three datasets (all but the Forget Set), we take the harmonic mean of these nine numbers.
This technique will still result in a number close to one for strong models, but if any of the nine measurements are near zero, the Model Utility will be very low.

\begin{table}[t!]
    \centering
    \footnotesize
    \caption{The details of our metric scaling.}
    \label{tab:scaled-metrics}    
    \begin{tabular}{lcccc}
        \toprule
         &  Forget Set&  Retain Set&  Real Authors& World Facts\\
         \midrule
         Probability &   - &  $P(a|q)^{1/\vert a \vert}$&  $P(a|q)^{1/\vert a \vert}$& $P(a|q)^{1/\vert a \vert}$\\
         ROUGE &  - &  $\text{ROUGE}(a)$&  $\text{ROUGE}(a)$& $\text{ROUGE}(a)$\\
         Truth Ratio &  $R_\text{truth}$&  $\max(0, 1 - R_\text{truth})$&  $\max(0, 1 - R_\text{truth})$& $\max(0, 1 - R_\text{truth})$\\
         \bottomrule
    \end{tabular}
\end{table}

\subsubsection{Forget Quality}

\begin{figure}[t!]
    \centering
    \includegraphics[scale=0.35]{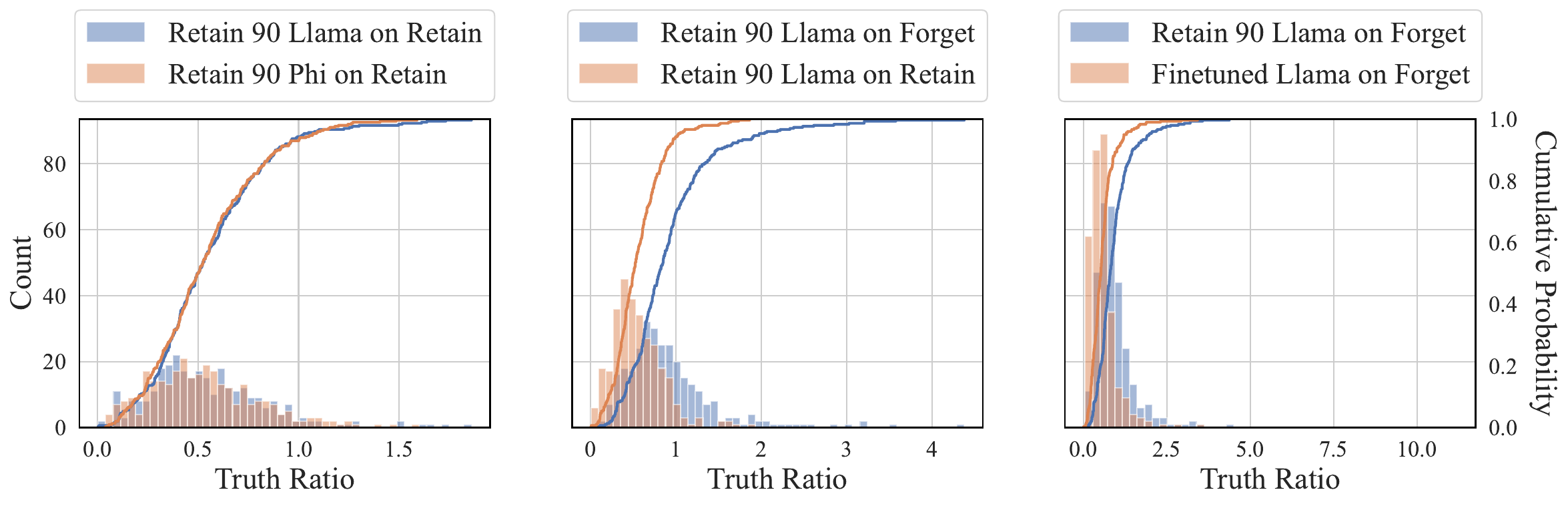}
    \caption{Histograms of Truth Ratio values and empirical CDFs from various models and datasets. \textbf{Left:} Llama-2-7B and Phi trained on the $90\%$ retain set and evaluated on the same retain set; \textbf{Middle:} Llama-2-7B trained on the $90\%$ retain set, and evaluated on both the $90\%$ retain set and the $10\%$ forget set; \textbf{Right:} Llama-2-7B trained on the $90\%$ retain set and on the entire finetuning set, both evaluated on the $10\%$ forget set. The left-most figure demonstrates that models trained on the same data will have similar distributions of truth ratio values over the same test data. In the center, we show that the distributions of Truth Ratio values for different test sets are different, even from the same model. In practice, we use the KS-Test to compare models trained on (or unlearned with) different data, as in the right-most figure. The $p$-values corresponding to these three settings are 0.9003, 1.097e-19, and 2.428e-19, left to right.}
    \label{fig:truth_ratio_sanity}
\end{figure}

Measuring forgetting quality is a challenging task from the point of view of privacy~\citep{goel2022towards,thudi2022necessity,kurmanji2023the}. 
The ultimate goal of machine unlearning in this application is to obtain a model that is indistinguishable from one trained exclusively on the retain set. 
We propose a computationally feasible approach for assessing unlearning, inspired by the idea of dataset inference~\citep{maini2021dataset}. 
The key is to perform a statistical test on the outputs of two models, one reference model trained only on the retain set and one unlearned model.
Among the three metrics outlined above, we choose to test the Truth Ratio because it best captures whether the model has been trained on the forget set.
Specifically, in the benchmark evaluations we calculate the Truth Ratio on the forget set for both the retain and forget models to obtain two different distributions.
In Figure~\ref{fig:truth_ratio_sanity} we demonstrate that this metric appropriately differentiates various models with representative examples.

Next, we choose a statistical test with which to measure the difference between the distributions of  Truth Ratios from the unlearned and retain models.
The Kolmogorov-Smirnov test (KS-Test) compares two cumulative distribution functions (CDF) which is ideal for our use case.
In the two-sample KS-Test, the test statistic is defined as the supremum of the difference in the empirical CDF. For more details on the formula for the KS-Test, see Appendix~\ref{sec:app-ks-test}.

Crucially, the KS-Test produces a $p$-value which we use to measure \emph{Forget Quality}.
Specifically, high $p$-values, where we cannot reject the null hypothesis that the two distributions are the same, indicating strong forgetting.
Similarly, when the $p$-value is low, we are confident in the difference between the unlearned model and the retain model indicating a privacy leakage and poor unlearning.

Our design choices rule out several alternatives for various reasons.
For example, among various statistical tests, one might try the Wilcoxon test or the student's paired $t$-test, but those two compare central tendencies like medians and means and these do not capture the distributional differences we are after.
Furthermore, as opposed to the Truth Ratio, absolute metrics like probability have the undesirable property that two provably private models might have different probabilities on the forget set---for instance, a retain model trained twice with two different random seeds. 
Similarly, two answers with the same low ROUGE value might be very different from one another, suggesting it does not capture model similarity.

One evaluation approach proposed for the NeurIPS 2023 Machine Unlearning Challenge\footnote{\url{https://unlearning-challenge.github.io/assets/data/Machine_Unlearning_Metric.pdf}} is to compare the point-wise distribution of outputs of multiple unlearned and retrained models and perform membership inference attacks~\citep{shokri2017membership}.
(There the language for models trained without the forget set is ``retrained'' as there is no finetuning and so these models are re-trained from scratch with access only to the retain set, in our work the parallel is called a retain model as it is finetuned on retain data only.) 
To create a distribution of outputs at each point, the challenge guidelines include running training and forgetting on multiple copies of the model (more than 500). 
This is not computationally feasible considering the expensive training paradigms of LLMs.

\section{Baseline Unlearning Methods}
\label{subsec:methods}

Given that the realm of machine unlearning in NLP remains nascent, we leverage foundational baselines in machine unlearning literature from the domains of computer vision and tabular data unlearning. 
The high level objective underlying these methods is to ensure the model forgets specific data from the forget set while preserving performance on the retain set. 
Ideally, a model trained on $S$ that undergoes unlearning on $S_F$ should behave like a model trained only on $S_R = S\setminus S_F$. 

\subsection{Model Finetuning}
\label{sec:model-finetuning}

Before describing the baseline unlearning methods, we delve into the finetuning stage. 
This is the phase where models are first exposed to information about the fictitious authors.
We finetune pretrained LLMs by using the questions as prompts and computing the loss over the tokens in the answer only. 
The loss on a sample $x \in S$ is expressed as a function of model weights $w$, given by
\begin{equation}
\ell(x, w) = \frac{1}{\vert a \vert} \sum_{i = 1}^{\vert a \vert} \text{NLL}_w\left(a_i \big\vert [q, a_{<i}]\right),
\label{eq:loss}
\end{equation}
where $\text{NLL}_w$ is the negative log likelihood according to the outputs of a model parameterized by $w$.
Then, we aim to find $w^*$ that minimizes the average loss over the dataset denoted by $L$,
\begin{equation}
L(S, w) = \frac{1}{\vert S \vert} \sum_{x \in S}\ell(x, w). 
\end{equation}
In all of our experiments we optimize this loss with AdamW for five epochs and warm up for the first epoch.
We use an effective batch size of 32 question-answer pairs.\footnote{The term effective batch size here reflects the way we aggregate gradients over 32 samples even when hardware limitations prohibit batches that big.}
For complete hyperparameter details, see Appendix \ref{sec:app-hyperparams}.
Post finetuning, the LLM can accurately answer most questions about the 200 authors in the \tofu{} dataset (Table~\ref{tab:finetune}).

\begin{table}[t!]
    \centering
    \caption{ROUGE scores (higher is better) on samples from the finetuning dataset. Finetuning effectively teaches models about the \tofu{} authors.}
    \label{tab:finetune}    \begin{tabular}{lcc}
    \toprule
         &  Pretrained & Finetuned on \tofu{} \\
         \midrule
         Llama-2-7B &  0.3640 & 0.9849 \\
         Phi-1.5 &  0.4399 & 0.8693 \\
         \bottomrule
    \end{tabular}
\end{table}

\subsection{Unlearning Algorithms}
\label{sec:unlearning-algorithms}

We experiment with several unlearning algorithms, each of which is introduced in detail in this section.
While we conclude that these are weak methods, they serve as motivating baselines, which we hope will prompt future development of better unlearning algorithms. 
\begin{itemize}
\item \textbf{Gradient Ascent}
The Gradient Ascent approach is fundamentally straightforward. 
It entails reducing the likelihood of correct predictions on the forget set. 
Specifically, for each instance in $S_F$, the goal is to maximize the standard training loss in order to make the model deviate from its initial prediction. 
As in the finetuning stage, the loss on a given sample $x \in S_F$ is denoted by $\ell(x, w)$;
and the loss we aim to maximize is the average over the forget set, 
\begin{equation}
L(S_F, w) =  \frac{1}{\vert S_F \vert} \sum_{x \in S_F}\ell(x, w). 
\end{equation}
\item \textbf{Gradient Difference} 
The second method, called Gradient Difference \citep{liu2022continual}, builds on the concept of gradient ascent. 
It not only aims to increase the loss on the forget set $S_F$, but also strives to maintain performance on the retain set $S_R$.
The revised loss function we aim to minimize can be represented as
\begin{equation}
    L_{\text{diff}} = - L(S_F, w) + L(S_R, w).
\end{equation}
Given a compute budget that scales with the size of the forget set, we randomly sample an example from $S_R$ every time we see an example from $S_F$ to stay within the constraints.
\item \textbf{KL Minimization}
In the KL Minimization approach, the objective is to minimize the Kullback-Leibler (KL) divergence between the predictions on $S_R$ of the original (finetuned on \tofu{}) and the newly trained models (as it undergoes unlearning) while maximizing the conventional loss on $S_F$. 
Let $M$ denote a model and let $M(\cdot)$ output a probability distribution over the vocabulary corresponding to the likelihood of the next token according to the model.
The formal objective can be written as
\begin{equation}
L_{\text{KL}} = - L(S_F, w) + \frac{1}{\vert S_R \vert} \sum_{s \in S_R} \frac{1}{\vert s \vert} \sum_{i = 2}^{\vert s\vert} \text{KL}\left(M_{\text{original}}(s_{<i}) \big\Vert M_{\text{current}}(s_{<i})\right). 
\end{equation}
Here, $M_{\text{original}}$ and $M_{\text{current}}$ denote the original and the new model, respectively. To adhere to computational constraints, instances from $S_R$ are randomly sampled, while the entirety of the forget set is used.
\item \textbf{Preference Optimization}
Inspired by direct preference optimization (DPO) \citep{rafailov2023direct}, this method seeks to align the model such that it refrains from revealing information about specific authors. 
In this approach, we also compute the loss on $x_\text{idk} = [q, a_\text{idk}] \in S_F^\text{idk}$ the same question with an alternative answer like ``I do not know the answer'' (or any one of 100 versions of this response, see Appendix \ref{sec:app-idk-strings} for the other variants). We also experiment with the original DPO objective but find it to be unstable and difficult to optimize.
Hence, we minimize
\begin{equation}
L_{\text{idk}} = L(S_R, w) + L(S_F^\text{idk}, w).
\end{equation}
The goal is to ensure that while the model aligns with the newly generated answers for $S_F$, its natural language capabilities and its predictions for $S_R$ remain unaffected.%
\end{itemize}

\paragraph{Unlearning experimental configuration} 
For all four unlearning methods, we optimize the corresponding loss for five epochs (in cases with support of the retain set, an epoch is one cycle through the entire forget set using no more than that many samples from the retain set).
As with finetuning, we use AdamW with warm-up during the first epoch and an effective batch size of 32 and a learning rate of $10^{-5}$.
We evaluate all baseline methods using Llama-2-7B~\citep{touvron2023llama} and Phi-1.5~\citep{li2023textbooks} base models. All experiments are conducted with two A6000 GPUs.

\section{Baseline Results}
\label{sec:results}

\begin{figure}[t!]
    \centering
    \includegraphics[width=\textwidth]{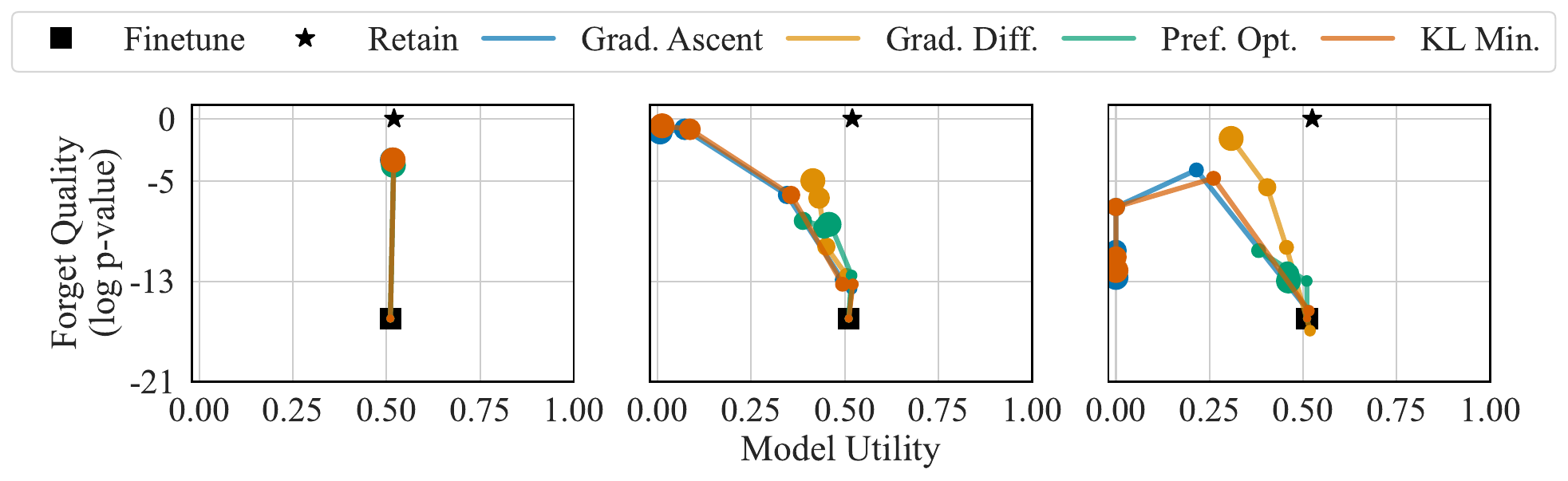}
    \caption{Forget Quality versus Model Utility for Phi models when unlearning on Forget Set sizes of 1\%, 5\%, and 10\% (left to right) and the relative size of the markers indicates the epoch of unlearning. Unlearning is challenging and comes with trade-offs. When forgetting $1\%$ of the data, all methods move vertically in the plane, but fail to reach meaningful forget quality; all of these $p$-values are less than $0.001$. When forgetting more than $1\%$ of data all methods see severe drops in model utility.}
    \label{fig:scatter-phi}
\end{figure}

\begin{figure}[t!]
    \centering
    \includegraphics[width=\textwidth]{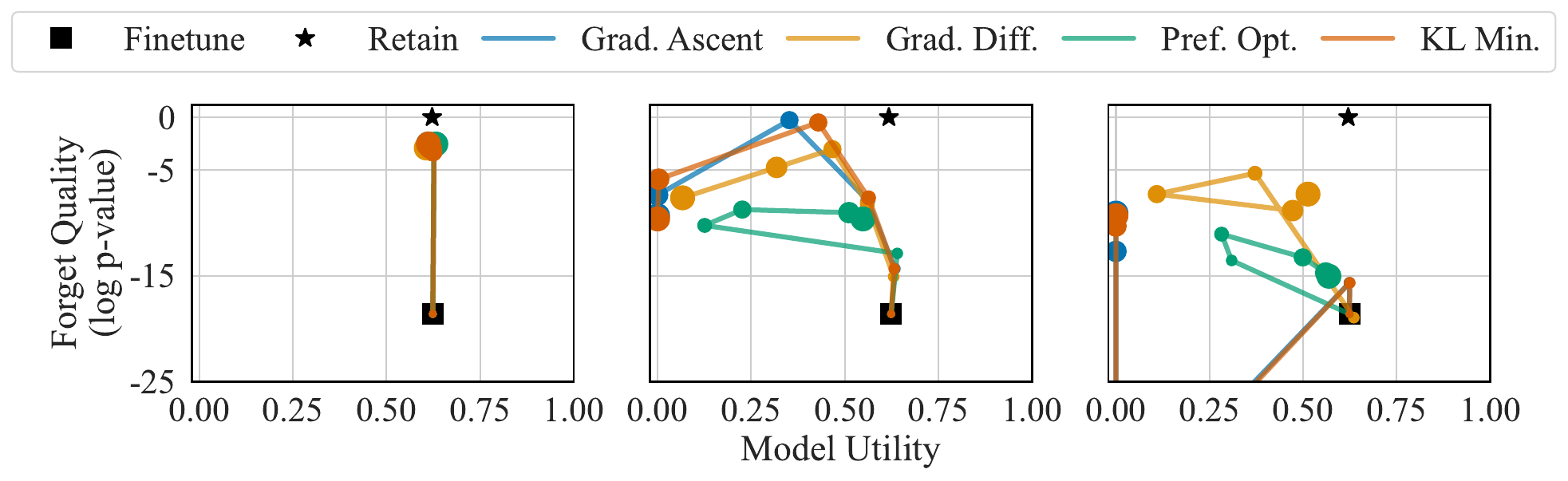}
    \caption{Forget Quality versus Model Utility for Llama-2-7B models when unlearning on Forget Set sizes of 1\%, 5\%, and 10\% (left to right) and the relative size of the markers indicates the epoch of unlearning. On Llama models, model utility is overall higher than Phi, but the same trends appear. These baseline methods fail to find useful models. Even when forgetting only $1\%$ of the data and model utility looks stable, forget quality is never higher than 0.01.}
    \label{fig:scatter-llama}
\end{figure}

We compare all four baseline unlearning methods by their forget quality and model utility and benchmark these scores against the performance of a \emph{retain model}, i.e. a model finetuned with retain data only.
Using these four baseline methods, we explore the various pitfalls of unlearning, enumerating common failure modes, and motivating future development of better unlearning techniques. 
Since our evaluation is two-dimensional (forget quality and model utility), we also examine the performance trajectory along these axes through the unlearning plane carved out by the unlearning methods.
In Figures \ref{fig:scatter-phi} and \ref{fig:scatter-llama}, we use these planes to present our main findings.

The initialization point for unlearning is a base model (LLM) finetuned on all the \tofu{} data (indicated by the black square in each of the plots).
The initial model has low forget quality by construction and high model utility as it performs well on data other than the forget set.
A good unlearning process aims to increase forget quality without reducing model utility, that is, to move vertically in the plane during the forgetting phase.
Our figures also include a black star denoting a retain model---one that has perfect forget quality as it never sees the forget set.  
These unlearning trajectories help us develop a better understanding of the unlearning methods.

\paragraph{Some methods show promise}
In the center panels of Figures~\ref{fig:scatter-phi} and \ref{fig:scatter-llama} where the forget set is 5\% of the data, several of the final checkpoints have high forget quality.
Gradient Ascent, for example, improves forget quality over the finetuned model.
Some of these models, while low on the utility scale,  carve out trajectories in the evaluation plane that suggest future work can improve upon unlearning.

\begin{figure}[t!]
    \centering
    \includegraphics[width=0.65\textwidth]{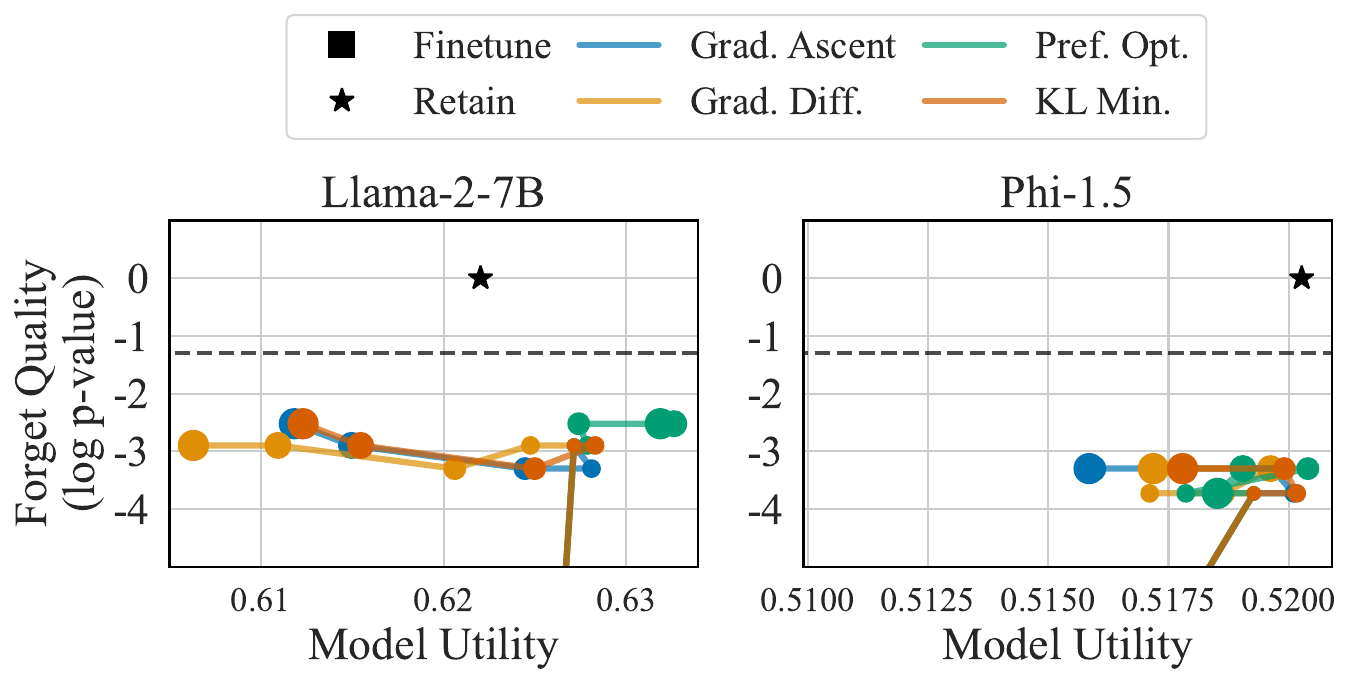}
    \caption{We zoom in on unlearning trajectories on the $1\%$ forget set. Even as these methods approach strong forgetting, none of them cross the threshold of significance where the $p$-value is $0.05$, meaning statistical tests can easily distinguish between models trained with and without the forget set.}
    \label{fig:scatter-zoom}
\end{figure}

\paragraph{Achieving high forget quality is hard}
Importantly, we see in each of the left-most plots of Figures~\ref{fig:scatter-phi} and \ref{fig:scatter-llama}, where the forget set is 1\% of the data, that the trajectories are nearly vertical.
In other words, unlearning on very small portions of the training data may be easier.
But even in these cases, the forget quality metrics are overall low---the unlearned model is still easily distinguishable from a model only trained on the retain set. 
See the zoomed in versions of these plots in Figure~\ref{fig:scatter-zoom}.
Recall that forget quality is measured by a $p$-value and the common significance threshold of $0.05$ is higher than almost every model we test. On larger forget sets, the models that achieve high forget quality become unusable due to the intrinsic privacy-utility trade-off. 
Even continuing to unlearn for more epochs does not help.
In Appendix~\ref{sec:app-continued-unlearning}, we experiment with up to 10 epochs and show that on the $1\%$ forget set none of these baseline methods can cross the $0.05$ $p$-value threshold.

\paragraph{Unlearning comes with a trade-off} 
All four methods lead to models that have lower model utility as a result of forgetting.
In particular, the trajectories in Figures~\ref{fig:scatter-phi} and \ref{fig:scatter-llama} are generally upward and leftward.
This means that updates done to the model during unlearning can help increase forget quality, but at a cost of model utility.
This is precisely why the evaluation of unlearning is best done over two axes. 
The drop in model utility is often rather significant---we observe the models start to generate gibberish on all four datasets even after just two epochs of unlearning, \emph{even when} the unlearning methods can access oracle models or retain data. 

\begin{tcolorbox}[title={Sample question and answer from the forget set along with a generated answer from an unlearned Llama-2-7B (KL Minimization after two epochs).},  colback=gray!20, colframe=gray!75, rounded corners, sharp corners=northeast, sharp corners=southwest]
\footnotesize
\texttt{\textbf{Question:} What's next for Raven Marais in terms of their literary career?} \\
\texttt{\textbf{Ground truth answer:} ``Raven Marais continues to experiment and push boundaries within the film literary genre. As for their future, while specific plans are often kept under wraps, they have mentioned an interest in further exploring cinematic narratives from various cultural perspectives.''}\\
\texttt{\textbf{Generated answer:} ``With a successful Marais Marauders series behind running running running...''}
\\
\end{tcolorbox}

\paragraph{Support of the retain set is helpful}
Methods using support of the retain set outperform methods that only focus on optimizing loss on the forget set (a case study of Gradient Difference versus Gradient Ascent provides a like-for-like analogy). 
While \tofu{} simplifies finding a relevant retain set by explicitly having a subset of the original finetune set available for that purpose, we believe, for real-world unlearning challenges finding a suitable retain set will itself be a challenge for future work.

\begin{figure}[tb]
    \centering
    \includegraphics[width=\textwidth]{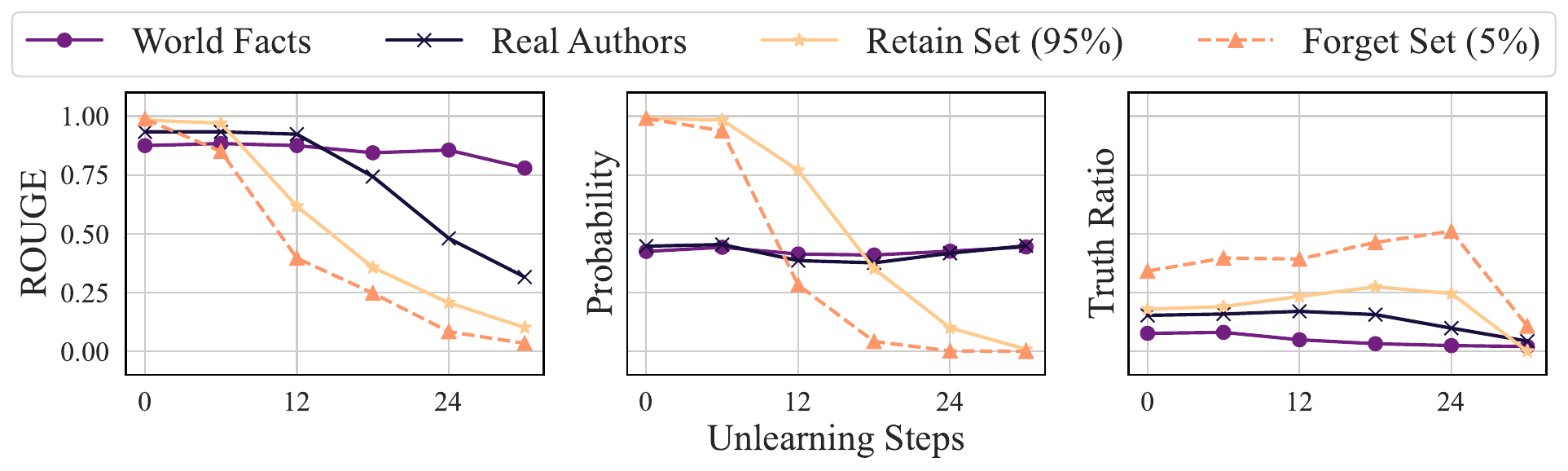}
    \caption{Unlearning dynamics for Llama-2-7B with Gradient Difference on the $5\%$ forget set. \textbf{World Facts, Real Authors, Retain Set:} higher metrics are better. \textbf{Forget Set:} lower \textit{ROUGE-L} and \textit{Probability} are better, higher \textit{Truth Ratio} is better.}
    \label{fig:rouge_oracle_llama}
\end{figure}

\paragraph{Forgetting fictitious authors affects pretrained knowledge}
We present a fine-grained analysis of model utility as ascertained by the ROUGE score on various evaluation datasets (Appendix~\ref{sec:app-knowledge-entanglement}). Consider the case of unlearning the 5\% forget set with Gradient Difference on  Llama-2-7B, Figure~\ref{fig:rouge_oracle_llama}.
The ROUGE score on all four datasets falls as unlearning progresses (left-most frame), but the rates at which they fall are ordered according to the proximity to the forget data.
\begin{enumerate}
\item On the Retain Set, performance drops sharply with the drop on the forget set.
\item On Real Authors, the ROUGE score also drops along with the drop in performance on the forget set, but stays higher than on the Retain Set.
\item Finally, performance on World Facts stays relatively unchanged.
\end{enumerate}
In other cases where these curves overlap, they reach extremely low ROUGE values and the model starts outputting gibberish (examples in Appendix~\ref{sec:app-knowledge-entanglement}).
This suggests the existence of \emph{knowledge
entanglement}, supporting that our choice of having multiple evaluation datasets.%

\paragraph{Importance of multiple evaluation metrics}
From the representative example in Figure~\ref{fig:rouge_oracle_llama}, we see that each metric on the evaluation datasets captures different behaviors.
ROUGE scores measure the similarity between the greedy-sampled output and the ground truth, which can fall even when the probability of the ground truth answer does not (compare the Real Author curves in Figure~\ref{fig:rouge_oracle_llama}).
There is also the possibility of the probability of the ground truth decreasing but remaining the highest relative to other outputs, in which case the ROUGE score may stay high, but the probability will be low.
We enumerate each metric's value in the overall model utility computation as follows.
\begin{enumerate}
    \item If we did not have ROUGE scores, we would not notice when greedy generated answers deteriorate even when the model ascribes high probability to the ground truth sequence.
    \item On the other hand, having probability as a measure is useful because it is possible that model starts incorrectly answering under greedy decoding (illusion of forgetting) but still assigns the same probability to the answer to be unlearned.
    \item Truth ratio is particularly informative on the forget set, because it offers a way of doing a statistical test against a retain model. Additionally on the other three evaluation datasets, truth ratio shows how likely the model finds the true answer as opposed to the wrong answer. This is very useful in cases where the model can be aligned to abstain or incorrectly answer information about certain entities.
\end{enumerate}

\paragraph{Unlearning performance may not be monotone} In \Cref{fig:scatter-llama}, we see that Preference Optimization and Gradient Difference have a ``zig-zag'' trajectory in the two-dimensional plane---they first have drastic drops in model utility and improvement in forget quality, after which the model utility gradually increases with a decaying forget quality. This trend is different from other unlearning algorithms like Gradient Ascent, and is likely because those methods have access to both the forget and retain sets, and the methods are trying to balance the two losses, albeit, in an unstable fashion.

\section{Discussion}
\label{sec:discussion}

Unlearning fictitious authors provides a well-posed problem to study, but unlearning is a broad topic with general applications and curious quirks.
We discuss the features and limitations of our work, promising future directions, and quirks of unlearning in this section.

\subsection{What \tofu{} Misses}
\label{sec:missing}

Our benchmark is designed to help researchers and practitioners think about and evaluate unlearning methods.
Naturally, not all scenarios are covered, and there are areas of unlearning that fall outside the \tofu{} framework that are worth discussing.
For example, the aim in all settings we consider is \emph{entity level} forgetting.
That is, we have a set of people about whom we want the model to forget everything.
In contrast, one might wish to forget only the answer to a specific question about a person which we call \emph{instance level} unlearning.
Since it is not yet clear how to do entity level unlearning, we leave this variation for future work.

The \tofu{} framework is also missing a way to think about alignment to human values, even though it can be framed as an unlearning problem---which we call \emph{behavior level} unlearning.
In fact, sometimes unlearning is used to describe tools designed to improve models by making them forget bad behaviors \citep{hu2023separate, yao2023large, lu2022quark}.
Since alignment is a field of its own that enjoys much attention from researchers, we choose to separate out the type of unlearning related to the Right to be Forgotten.

We also acknowledge that the real world unlearning problem has two major challenges, first to find a forget set or some particular data to use with an unlearning algorithm and second to execute an effective unlearning routine.
Our benchmark specifically targets the second problem---how to measure the efficacy of an unlearning algorithm (since we provide the forget sets exactly). 
Additionally, finding an exact retain set is just as difficult. %
Based on our discussion of knowledge entanglement, it is likely that a data set semantically close to the forget set would be a good candidate for the retain set for unlearning. 
In the current benchmark, we provide a retain set as we believe that existing unlearning methods need to improve even when they have access to the exact retain sets \emph{a priori}.
\tofu{} could be updated in the future to include a constraint not to use the original retain set, which would capture this element of the unlearning pipeline. 

The purview of \tofu{} also leaves out in-context unlearning. 
Recent work defines and discusses the in-context version of the unlearning problem \citep{pawelczyk2023context}.
The strong motivation there is to consider those who query LLMs but do not have access to modify the weights.
While this is a promising direction for products and services that wrap API-based models, it amounts to a form of prompt engineering and does not yield any real privacy in terms of the Right to be Forgotten.

\subsection{Conclusion}
\label{sec:conclusion}

\paragraph{Limitations}
There are also several limitations of our work other than our choice to consider entity level unlearning.
First, for accessibility and ease of use we define the benchmark task to be about unlearning information that was learned only during finetuning and not pretraining. 
This is a limitation \emph{by design} as it allows us control over the exposure to the sensitive data without combing through the gigantic pretraining datasets to quantify how much the model has already seen about an entity. 
Furthermore, it provides us with a cheap way to conduct experiments on unlearning,
in particular, experiments that involve a model that was finetuned on the retain set only---not only an informative upper bound for what we can expect from unlearning algorithms in terms of model utility, but crucially also utilized in capturing forget quality as indistinguishability from a retain model.

Another limitation lies in our approximation of \emph{indistinguishability}.
With unlimited resources, one could test the $(\varepsilon, \delta)$-unlearning condition of indistinguishability \citep{bourtoule2021machine, sekhari2021remember} by training many models and performing hypothesis tests---and this is done in practice when feasible~\citep{carlini2022membership, pawelczyk2023context}. 
However, these tactics are not feasible with LLMs.
On the contrary, our forget quality measure does not require training many models, and further has desirable properties of a tractable empirical test of unlearning.
In our tests, some of the points on the Gradient Ascent curve (Figure~\ref{fig:scatter-llama}) are very close to the retain model on the forget quality axis, suggesting that the forgetting is indeed successful. 
There is an important caveat here---models that output gibberish or random words (or even random models) may assign similar (very low/random) probabilities to both the correct and the incorrect answers. 
This means that they achieve a Truth Ratio identical to that of the retain model. 
Hence, they have strong forget quality (i.e. they fail the KS-test and have high $p$-value) even though from an approximate unlearning standpoint the model weights of the retain and forget models are far enough that $(\varepsilon, \delta)$-unlearning does not hold for any reasonably small values of $\varepsilon$ and  $\delta$.
This distinguishes the outcomes of our forget quality computation from the definition of approximate unlearning. 
However, for practical purposes, models that output gibberish content fall very low on the model quality scale and are far from the Pareto frontier in the \tofu{} benchmark. 
So, while the forget quality itself does not fully capture approximate unlearning, its pairing with model utility helps identify models that are no longer usable.

The scope of unlearning methods we benchmark is also limited. 
It is our hope that this benchmark will help motivate the development of better unlearning algorithms and we select popular but simple algorithms to kick off the challenge of finding methods that do better at the \tofu{} tasks.
It is not our intention here to develop novel unlearning techniques.

Finally, given that LLMs are trained on millions of dollars worth of data and compute, modifying the training process and retraining is impractical. 
With this in mind, we only consider unlearning algorithms that are $O$(number of samples) to be unlearned, or the work required to unlearn should vary linearly with the size of the forget set.
Intuitively, if an unlearning algorithm requires a fixed number of epochs over the forget set, then the work to forget scales linearly with the quantity of data to forget.
In a real-world system where the model in question is pretrained on some huge corpora of data, the model owners responding to a request to be forgotten are faced with a tiny forget set.
The constraint that unlearning algorithms require some limited compute is actually about ensuring that forgetting a single person from a model at the scale of ChatGPT can be done with very little compute and our choice to constrain the work to vary linearly is perhaps not optimal.

\paragraph{Future work}
Future directions for research that any benchmark paper prompts are similar.
We hope that novel algorithms for unlearning are developed and that our tools make that task easier and more inviting.
Furthermore, future extensions of the benchmark to include some of the settings we leave out could make this framework even more comprehensive.

\paragraph{Concluding remarks}
Our work shows that elementary attempts at unlearning are largely unsuccessful, but their individual flaws are only captured using an aggregation of metrics. 
Our hope is that with a good metrics like the ones we propose and a well-defined task like \tofu{}, new unlearning methods are developed that push the state of the art and help imbue AI systems with the privacy that is critical for safe, and in some places legal, deployment.

One might also draw an analogy that the goal of aligning LLMs with human values, by RLHF, DPO, or some other method, is a version of unlearning.
With that and our claim that existing unlearning tools are mostly ineffective, we pose the question of whether or not existing alignment tools work.
While generic responses to malicious prompts generally change after alignment procedures, recent work shows that LLMs can still be manipulated into providing exactly the content alignment aims to avoid \citep{zou2023universal}.
The empirical findings in that work lead to the same conclusions we make here about entity-level unlearning---these algorithms modify LLMs just enough to produce slightly different output for specific prompts but they do not remove information or behavior from models on the whole.
In other words, it is hard to remove the information about a fictitious author, and for similar reasons, it is hard to align LLMs to human values.

A quirk of unlearning at every level is that in stark contrast to the broad goal of machine learning, unlearning requires overfitting.
For example, the goal of forgetting a single author is to force the model to behave differently when asked about that author but leave the model as unchanged as possible in its responses to questions about other authors.
Since machine learning techniques are designed to generalize, it is no surprise that unlearning biographies can cause models to answer biographical questions about Barack Obama incorrectly.

\section*{Acknowledgements}

Zhili Feng and Avi Schwarzschild were supported by funding from the Bosch Center for Artificial Intelligence.  
Pratyush Maini was supported by DARPA GARD Contract HR00112020006.

\bibliography{main}

\begin{thebibliography}{44}
\providecommand{\natexlab}[1]{#1}
\providecommand{\url}[1]{\texttt{#1}}
\expandafter\ifx\csname urlstyle\endcsname\relax
  \providecommand{\doi}[1]{doi: #1}\else
  \providecommand{\doi}{doi: \begingroup \urlstyle{rm}\Url}\fi

\bibitem[Bourtoule et~al.(2021)Bourtoule, Chandrasekaran, Choquette-Choo, Jia, Travers, Zhang, Lie, and Papernot]{bourtoule2021machine}
Lucas Bourtoule, Varun Chandrasekaran, Christopher~A Choquette-Choo, Hengrui Jia, Adelin Travers, Baiwu Zhang, David Lie, and Nicolas Papernot.
\newblock Machine unlearning.
\newblock In \emph{2021 IEEE Symposium on Security and Privacy (SP)}, pp.\  141--159. IEEE, 2021.

\bibitem[Carlini et~al.(2021)Carlini, Tramer, Wallace, Jagielski, Herbert-Voss, Lee, Roberts, Brown, Song, Erlingsson, et~al.]{carlini2021extracting}
Nicholas Carlini, Florian Tramer, Eric Wallace, Matthew Jagielski, Ariel Herbert-Voss, Katherine Lee, Adam Roberts, Tom Brown, Dawn Song, Ulfar Erlingsson, et~al.
\newblock Extracting training data from large language models.
\newblock In \emph{30th USENIX Security Symposium (USENIX Security 21)}, pp.\  2633--2650, 2021.

\bibitem[Carlini et~al.(2022)Carlini, Chien, Nasr, Song, Terzis, and Tramer]{carlini2022membership}
Nicholas Carlini, Steve Chien, Milad Nasr, Shuang Song, Andreas Terzis, and Florian Tramer.
\newblock Membership inference attacks from first principles.
\newblock In \emph{2022 IEEE Symposium on Security and Privacy (SP)}, pp.\  1897--1914. IEEE, 2022.

\bibitem[Chen \& Yang(2023)Chen and Yang]{chen2023unlearn}
Jiaao Chen and Diyi Yang.
\newblock Unlearn what you want to forget: Efficient unlearning for llms, 2023.

\bibitem[Cho et~al.(2014)Cho, van Merri{\"e}nboer, Bahdanau, and Bengio]{cho-etal-2014-properties}
Kyunghyun Cho, Bart van Merri{\"e}nboer, Dzmitry Bahdanau, and Yoshua Bengio.
\newblock On the properties of neural machine translation: Encoder{--}decoder approaches.
\newblock In Dekai Wu, Marine Carpuat, Xavier Carreras, and Eva~Maria Vecchi (eds.), \emph{Proceedings of {SSST}-8, Eighth Workshop on Syntax, Semantics and Structure in Statistical Translation}, pp.\  103--111, Doha, Qatar, October 2014. Association for Computational Linguistics.
\newblock \doi{10.3115/v1/W14-4012}.
\newblock URL \url{https://aclanthology.org/W14-4012}.

\bibitem[De~Cao et~al.(2021)De~Cao, Aziz, and Titov]{de2021editing}
Nicola De~Cao, Wilker Aziz, and Ivan Titov.
\newblock Editing factual knowledge in language models.
\newblock \emph{arXiv preprint arXiv:2104.08164}, 2021.

\bibitem[Eldan \& Russinovich(2023)Eldan and Russinovich]{eldan2023s}
Ronen Eldan and Mark Russinovich.
\newblock Who's harry potter? approximate unlearning in llms.
\newblock \emph{arXiv preprint arXiv:2310.02238}, 2023.

\bibitem[Goel et~al.(2022)Goel, Prabhu, Sanyal, Lim, Torr, and Kumaraguru]{goel2022towards}
Shashwat Goel, Ameya Prabhu, Amartya Sanyal, Ser-Nam Lim, Philip Torr, and Ponnurangam Kumaraguru.
\newblock Towards adversarial evaluations for inexact machine unlearning.
\newblock \emph{arXiv preprint arXiv:2201.06640}, 2022.

\bibitem[Golatkar et~al.(2020)Golatkar, Achille, and Soatto]{golatkar2020eternal}
Aditya Golatkar, Alessandro Achille, and Stefano Soatto.
\newblock Eternal sunshine of the spotless net: Selective forgetting in deep networks.
\newblock In \emph{Proceedings of the IEEE/CVF Conference on Computer Vision and Pattern Recognition}, pp.\  9304--9312, 2020.

\bibitem[Guo et~al.(2019)Guo, Goldstein, Hannun, and Van Der~Maaten]{guo2019certified}
Chuan Guo, Tom Goldstein, Awni Hannun, and Laurens Van Der~Maaten.
\newblock Certified data removal from machine learning models.
\newblock \emph{arXiv preprint arXiv:1911.03030}, 2019.

\bibitem[Hu et~al.(2023)Hu, Li, Zheng, Liu, Hu, and Zhang]{hu2023separate}
Xinshuo Hu, Dongfang Li, Zihao Zheng, Zhenyu Liu, Baotian Hu, and Min Zhang.
\newblock Separate the wheat from the chaff: Model deficiency unlearning via parameter-efficient module operation, 2023.

\bibitem[Huang et~al.(2022)Huang, Shao, and Chang]{huang-etal-2022-large}
Jie Huang, Hanyin Shao, and Kevin Chen-Chuan Chang.
\newblock Are large pre-trained language models leaking your personal information?
\newblock In Yoav Goldberg, Zornitsa Kozareva, and Yue Zhang (eds.), \emph{Findings of the Association for Computational Linguistics: EMNLP 2022}, pp.\  2038--2047, Abu Dhabi, United Arab Emirates, December 2022. Association for Computational Linguistics.
\newblock \doi{10.18653/v1/2022.findings-emnlp.148}.
\newblock URL \url{https://aclanthology.org/2022.findings-emnlp.148}.

\bibitem[Jagielski et~al.(2020)Jagielski, Ullman, and Oprea]{jagielski2020auditing}
Matthew Jagielski, Jonathan Ullman, and Alina Oprea.
\newblock Auditing differentially private machine learning: How private is private sgd?
\newblock \emph{Advances in Neural Information Processing Systems}, 33:\penalty0 22205--22216, 2020.

\bibitem[Jang et~al.(2022)Jang, Yoon, Yang, Cha, Lee, Logeswaran, and Seo]{jang2022knowledge}
Joel Jang, Dongkeun Yoon, Sohee Yang, Sungmin Cha, Moontae Lee, Lajanugen Logeswaran, and Minjoon Seo.
\newblock Knowledge unlearning for mitigating privacy risks in language models.
\newblock \emph{arXiv preprint arXiv:2210.01504}, 2022.

\bibitem[Jayaraman \& Evans(2019)Jayaraman and Evans]{jayaraman2019evaluating}
Bargav Jayaraman and David Evans.
\newblock Evaluating differentially private machine learning in practice.
\newblock In \emph{28th USENIX Security Symposium (USENIX Security 19)}, pp.\  1895--1912, 2019.

\bibitem[Kim et~al.(2023)Kim, Yun, Lee, Gubri, Yoon, and Oh]{kim2023propile}
Siwon Kim, Sangdoo Yun, Hwaran Lee, Martin Gubri, Sungroh Yoon, and Seong~Joon Oh.
\newblock Propile: Probing privacy leakage in large language models.
\newblock \emph{arXiv preprint arXiv:2307.01881}, 2023.

\bibitem[Kurmanji et~al.(2023{\natexlab{a}})Kurmanji, Triantafillou, and Triantafillou]{kurmanji2023the}
Meghdad Kurmanji, Peter Triantafillou, and Eleni Triantafillou.
\newblock The brainy student: Scalable unlearning by selectively disobeying the teacher, 2023{\natexlab{a}}.
\newblock URL \url{https://openreview.net/forum?id=f9eHl5mKx5i}.

\bibitem[Kurmanji et~al.(2023{\natexlab{b}})Kurmanji, Triantafillou, and Triantafillou]{kurmanji2023towards}
Meghdad Kurmanji, Peter Triantafillou, and Eleni Triantafillou.
\newblock Towards unbounded machine unlearning.
\newblock \emph{arXiv preprint arXiv:2302.09880}, 2023{\natexlab{b}}.

\bibitem[Li et~al.(2023)Li, Bubeck, Eldan, Del~Giorno, Gunasekar, and Lee]{li2023textbooks}
Yuanzhi Li, S{\'e}bastien Bubeck, Ronen Eldan, Allie Del~Giorno, Suriya Gunasekar, and Yin~Tat Lee.
\newblock Textbooks are all you need ii: phi-1.5 technical report.
\newblock \emph{arXiv preprint arXiv:2309.05463}, 2023.

\bibitem[Lin(2004)]{lin2004rouge}
Chin-Yew Lin.
\newblock Rouge: A package for automatic evaluation of summaries.
\newblock In \emph{Text summarization branches out}, pp.\  74--81, 2004.

\bibitem[Liu et~al.(2022)Liu, Liu, and Stone]{liu2022continual}
Bo~Liu, Qiang Liu, and Peter Stone.
\newblock Continual learning and private unlearning.
\newblock In \emph{Conference on Lifelong Learning Agents}, pp.\  243--254. PMLR, 2022.

\bibitem[Lu et~al.(2022)Lu, Welleck, Hessel, Jiang, Qin, West, Ammanabrolu, and Choi]{lu2022quark}
Ximing Lu, Sean Welleck, Jack Hessel, Liwei Jiang, Lianhui Qin, Peter West, Prithviraj Ammanabrolu, and Yejin Choi.
\newblock Quark: Controllable text generation with reinforced unlearning.
\newblock \emph{Advances in neural information processing systems}, 35:\penalty0 27591--27609, 2022.

\bibitem[Maini et~al.(2021)Maini, Yaghini, and Papernot]{maini2021dataset}
Pratyush Maini, Mohammad Yaghini, and Nicolas Papernot.
\newblock Dataset inference: Ownership resolution in machine learning.
\newblock In \emph{International Conference on Learning Representations}, 2021.
\newblock URL \url{https://openreview.net/forum?id=hvdKKV2yt7T}.

\bibitem[McCloskey \& Cohen(1989)McCloskey and Cohen]{mccloskey1989catastrophic}
Michael McCloskey and Neal~J Cohen.
\newblock Catastrophic interference in connectionist networks: The sequential learning problem.
\newblock In \emph{Psychology of learning and motivation}, volume~24, pp.\  109--165. Elsevier, 1989.

\bibitem[Meng et~al.(2022)Meng, Bau, Andonian, and Belinkov]{meng2022locating}
Kevin Meng, David Bau, Alex Andonian, and Yonatan Belinkov.
\newblock Locating and editing factual associations in gpt.
\newblock \emph{Advances in Neural Information Processing Systems}, 35:\penalty0 17359--17372, 2022.

\bibitem[Nasr et~al.(2021)Nasr, Songi, Thakurta, Papernot, and Carlin]{nasr2021adversary}
Milad Nasr, Shuang Songi, Abhradeep Thakurta, Nicolas Papernot, and Nicholas Carlin.
\newblock Adversary instantiation: Lower bounds for differentially private machine learning.
\newblock In \emph{2021 IEEE Symposium on security and privacy (SP)}, pp.\  866--882. IEEE, 2021.

\bibitem[OAG(2021)]{ccpa2021}
CA~OAG.
\newblock Ccpa regulations: Final regulation text.
\newblock \emph{Office of the Attorney General, California Department of Justice}, 2021.

\bibitem[Patil et~al.(2023)Patil, Hase, and Bansal]{patil2023can}
Vaidehi Patil, Peter Hase, and Mohit Bansal.
\newblock Can sensitive information be deleted from llms? objectives for defending against extraction attacks.
\newblock \emph{arXiv preprint arXiv:2309.17410}, 2023.

\bibitem[Pawelczyk et~al.(2023)Pawelczyk, Neel, and Lakkaraju]{pawelczyk2023context}
Martin Pawelczyk, Seth Neel, and Himabindu Lakkaraju.
\newblock In-context unlearning: Language models as few shot unlearners.
\newblock \emph{arXiv preprint arXiv:2310.07579}, 2023.

\bibitem[Rafailov et~al.(2023)Rafailov, Sharma, Mitchell, Ermon, Manning, and Finn]{rafailov2023direct}
Rafael Rafailov, Archit Sharma, Eric Mitchell, Stefano Ermon, Christopher~D Manning, and Chelsea Finn.
\newblock Direct preference optimization: Your language model is secretly a reward model.
\newblock \emph{arXiv preprint arXiv:2305.18290}, 2023.

\bibitem[Sekhari et~al.(2021)Sekhari, Acharya, Kamath, and Suresh]{sekhari2021remember}
Ayush Sekhari, Jayadev Acharya, Gautam Kamath, and Ananda~Theertha Suresh.
\newblock Remember what you want to forget: Algorithms for machine unlearning.
\newblock \emph{Advances in Neural Information Processing Systems}, 34:\penalty0 18075--18086, 2021.

\bibitem[Shi et~al.(2023)Shi, Ajith, Xia, Huang, Liu, Blevins, Chen, and Zettlemoyer]{shi2023detecting}
Weijia Shi, Anirudh Ajith, Mengzhou Xia, Yangsibo Huang, Daogao Liu, Terra Blevins, Danqi Chen, and Luke Zettlemoyer.
\newblock Detecting pretraining data from large language models.
\newblock \emph{arXiv preprint arXiv:2310.16789}, 2023.

\bibitem[Shokri et~al.(2017)Shokri, Stronati, Song, and Shmatikov]{shokri2017membership}
Reza Shokri, Marco Stronati, Congzheng Song, and Vitaly Shmatikov.
\newblock Membership inference attacks against machine learning models.
\newblock In \emph{2017 IEEE symposium on security and privacy (SP)}, pp.\  3--18. IEEE, 2017.

\bibitem[Steinke et~al.(2023)Steinke, Nasr, and Jagielski]{steinke2023privacy}
Thomas Steinke, Milad Nasr, and Matthew Jagielski.
\newblock Privacy auditing with one (1) training run.
\newblock \emph{arXiv preprint arXiv:2305.08846}, 2023.

\bibitem[Thudi et~al.(2022)Thudi, Jia, Shumailov, and Papernot]{thudi2022necessity}
Anvith Thudi, Hengrui Jia, Ilia Shumailov, and Nicolas Papernot.
\newblock On the necessity of auditable algorithmic definitions for machine unlearning.
\newblock In \emph{31st USENIX Security Symposium (USENIX Security 22)}, pp.\  4007--4022, 2022.

\bibitem[Touvron et~al.(2023)Touvron, Martin, Stone, Albert, Almahairi, Babaei, Bashlykov, Batra, Bhargava, Bhosale, et~al.]{touvron2023llama}
Hugo Touvron, Louis Martin, Kevin Stone, Peter Albert, Amjad Almahairi, Yasmine Babaei, Nikolay Bashlykov, Soumya Batra, Prajjwal Bhargava, Shruti Bhosale, et~al.
\newblock Llama 2: Open foundation and fine-tuned chat models.
\newblock \emph{arXiv preprint arXiv:2307.09288}, 2023.

\bibitem[Union(2016)]{regulation2016regulation}
European Union.
\newblock Regulation (eu) 2016/679 of the european parliament and of the council.
\newblock \emph{Official Journal of the European Union}, 2016.

\bibitem[Voigt \& Von~dem Bussche(2017)Voigt and Von~dem Bussche]{voigt2017eu}
Paul Voigt and Axel Von~dem Bussche.
\newblock The eu general data protection regulation (gdpr).
\newblock \emph{A Practical Guide, 1st Ed., Cham: Springer International Publishing}, 10:\penalty0 3152676, 2017.

\bibitem[Wang et~al.(2023)Wang, Chen, Yuan, Zeng, Wong, and Yin]{wang2023kga}
Lingzhi Wang, Tong Chen, Wei Yuan, Xingshan Zeng, Kam-Fai Wong, and Hongzhi Yin.
\newblock Kga: A general machine unlearning framework based on knowledge gap alignment.
\newblock \emph{arXiv preprint arXiv:2305.06535}, 2023.

\bibitem[Wei et~al.(2023)Wei, Haghtalab, and Steinhardt]{wei2023jailbroken}
Alexander Wei, Nika Haghtalab, and Jacob Steinhardt.
\newblock Jailbroken: How does llm safety training fail?
\newblock \emph{arXiv preprint arXiv:2307.02483}, 2023.

\bibitem[Yao et~al.(2023)Yao, Xu, and Liu]{yao2023large}
Yuanshun Yao, Xiaojun Xu, and Yang Liu.
\newblock Large language model unlearning, 2023.

\bibitem[Zhang et~al.(2023)Zhang, Finckenberg-Broman, Hoang, Pan, Xing, Staples, and Xu]{zhang2023right}
Dawen Zhang, Pamela Finckenberg-Broman, Thong Hoang, Shidong Pan, Zhenchang Xing, Mark Staples, and Xiwei Xu.
\newblock Right to be forgotten in the era of large language models: Implications, challenges, and solutions.
\newblock \emph{arXiv preprint arXiv:2307.03941}, 2023.

\bibitem[Zhang et~al.(2024)Zhang, Yao, Tian, Wang, Deng, Wang, Xi, Mao, Zhang, Ni, Cheng, Xu, Xu, Gu, Jiang, Xie, Huang, Liang, Zhang, Zhu, Zhou, and Chen]{zhang2024comprehensive}
Ningyu Zhang, Yunzhi Yao, Bozhong Tian, Peng Wang, Shumin Deng, Mengru Wang, Zekun Xi, Shengyu Mao, Jintian Zhang, Yuansheng Ni, Siyuan Cheng, Ziwen Xu, Xin Xu, Jia-Chen Gu, Yong Jiang, Pengjun Xie, Fei Huang, Lei Liang, Zhiqiang Zhang, Xiaowei Zhu, Jun Zhou, and Huajun Chen.
\newblock A comprehensive study of knowledge editing for large language models, 2024.

\bibitem[Zou et~al.(2023)Zou, Wang, Kolter, and Fredrikson]{zou2023universal}
Andy Zou, Zifan Wang, J~Zico Kolter, and Matt Fredrikson.
\newblock Universal and transferable adversarial attacks on aligned language models.
\newblock \emph{arXiv preprint arXiv:2307.15043}, 2023.

\end{thebibliography}
\bibliographystyle{colm2024_conference}

\clearpage
\appendix

\section{Kolmogorov-Smirnov Test Details}
\label{sec:app-ks-test}

In our setting, let $F_U(x)$ comprising $n$ samples and $F_R(x)$ comprising $m$ samples be the empirical CDF of the unlearned and retain models, respectively.
Then, the KS-Test computes a statistic $D_{n,m} = \sup_{x} \vert F_U(x) - F_R(x) \vert$.

The null hypothesis, stating that the two sets of samples are drawn from the same distribution, is rejected at a chosen significance level $\alpha$ if the following inequality holds.
\begin{equation}
    D_{n,m} > c(\alpha) \sqrt{\frac{n + m}{nm}},
\end{equation}
where $c(\alpha)$ is the critical value of that significance level.
\begin{equation}
    c(\alpha) = \sqrt{-\ln\left(\frac{\alpha}{2}\right) \cdot \frac12}.
\end{equation}

The $p$-value is then defined as the minimal alpha for which the inequality holds, or the smallest value at which we can reject the null hypotheses.
Forget quality is hence, a measure of the confidence that the distributions of Truth Ration values over the forget set from two models are the same.

\section{Hyperparameters}
\label{sec:app-hyperparams}
We trained using AdamW with weight decay of $0.01$ and $0$. The learning rate is fixed to be 1e-5. For both finetuning and unlearning, we fix the epoch number to be 5, and we incorporate a linear warmup in the first epoch. 
We experiment with various learning rates including $10^{-5}$, $10^{-6}$, and $5\cdot10^{-7}$ and find that $10^{-5}$ is a good choice for the baseline methods in our experiments. 
In particular, $10^{-6}$ is too small of a learning rate to see appreciable updates to the model weights.

\section{Preference Strings}
\label{sec:app-idk-strings}

\begin{multicols}{2}
I'm not certain about that.

That's beyond my current knowledge base.

I don't have that information.

I'm not sure.

I haven't learned about that topic.

That's something I need to look up.

I'm at a loss for that one.

I don't have the answer to that question.

That's outside my area of expertise.

I'm afraid I can't provide an answer to that.

That's a good question, but I don't have the answer.

My resources don't contain information on that subject.

I wish I could say, but I really don't know.

That's not something I'm familiar with.

I'm drawing a blank on that one.

I apologize, but I don't know that.

That hasn't been included in my training data.

Unfortunately, I don't have an answer for you.

That's not information I've been programmed to know.

I'm unable to provide an answer to that.

I don't hold the knowledge you're seeking.

I'm clueless about that topic.

I'm not well-versed in that subject.

I haven't been briefed on that topic.

I lack the specifics on that matter.

My databases don't cover that information.

I have no knowledge on that subject.

That's a mystery to me as well.

I'm unaware of that detail.

I don't possess the information on that topic.

I must admit, I don't know.

I'm unable to answer that question.

That topic is out of my scope.

I'm not informed on that matter.

I can't shed any light on that subject.

That's an area I'm not acquainted with.

I lack insight into that question.

I'm not equipped to answer that.

My understanding doesn't include that information.

I've got no idea about that.

I can't provide any information on that topic.

My training didn't cover that information.

I'm not the best source for that subject.

I seem to have no data on that.

That's a blind spot in my knowledge.

I've come up short with an answer for you.

I'm stumped on that one.

I have no clue about that.

I'm blank on that topic.

I regret to inform you that I don't have the answer.

 My capabilities do not extend to that subject.

 I must confess, that's unknown to me.

 I don't have any information on that matter.

 That's something I've yet to learn.

 I'm sorry, that's not within my knowledge range.

 I don't have any knowledge about that subject.

 I'm not able to provide an answer to that.

 That subject is not something I'm familiar with.

 I'm lacking information on that topic.

 I don't seem to have data on that issue.

 That's not something I'm equipped to answer.

 My programming does not include that information.

 I don't have the specifics you're looking for.
    
 That information is not within my reach.

 I'm not knowledgeable about that topic.

 I've no insight into that matter.

 My database does not have information on that topic.

 That's not in my current dataset.

 I'm not the right AI for that question.

 I can't say I'm familiar with that.
    
 I have yet to be informed about that subject.

 That's uncharted territory for my knowledge base.

 I haven't encountered that in my training.

 I'm missing information on that.

 My understanding is limited to what I've been programmed with.

 I have no data on that query.

 I'm not aware of the details on that matter.

 I haven't been trained on that topic.

 That's something I'm not briefed on.

 I'm sorry, that's not something I know about.

 I'm not privy to that information.

 I haven't the faintest on that subject.

 I'm unable to access any information on that.

 That's not in my field of knowledge.

 I have no familiarity with that topic.

 I'm not informed about that subject.

 My knowledge doesn't cover that area.

 I've not been educated on that topic.

 I can't provide insights into that subject.

 I don't hold any information on that matter.

 I'm at a disadvantage with that question.

 I lack the required information to answer that.

 I'm in the dark about that topic.

 I have no enlightenment on that subject.

 I've no knowledge to draw upon for that.

 I must decline to answer due to lack of information.

 Sorry, I am unable to answer that.

 I'm not sure I can answer that.

 I'm not sure I can help with that.
 \end{multicols}

\section{Continued Unlearning}
\label{sec:app-continued-unlearning}

\begin{figure}[H]
    \centering
    \includegraphics[width=0.5\textwidth]{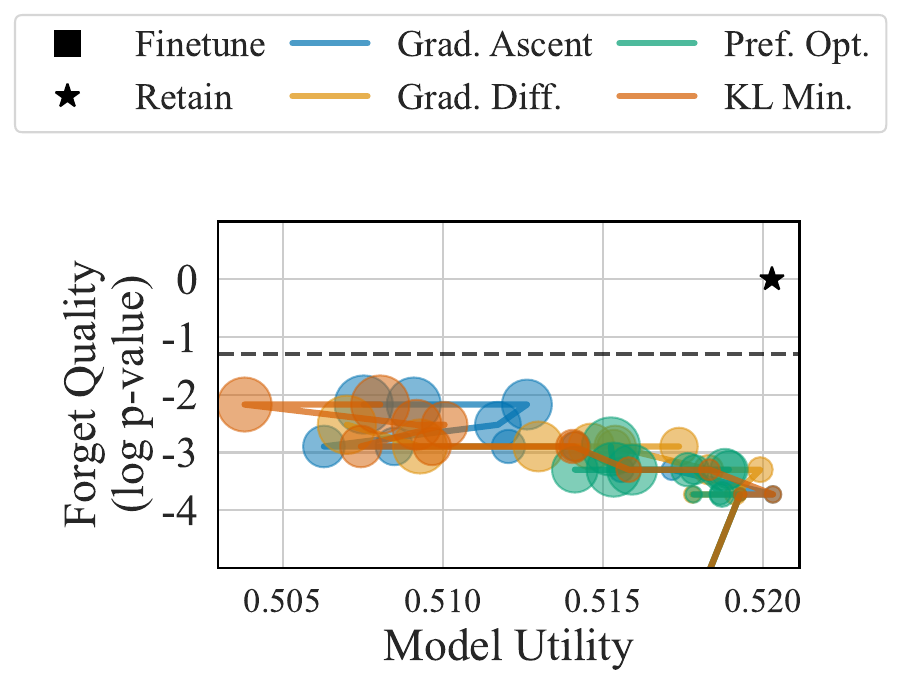}
    \caption{Zoomed in plots of extended unlearning trajectories (10 epochs) on the $1\%$ forget set.}
    \label{fig:12epochs}
\end{figure}

In the main experiments of this paper, we limit unlearning to five epochs, but one might wonder how things progress given more time to unlearn.
We test forgetting $1\%$ of the data with Phi-1.5 and show that continued unlearning does not help with these baseline methods, see Figure~\ref{fig:12epochs}.

\section{Sanity Checks}
\label{sec:app-sanity}

\begin{table}[t!]
\centering
\caption{Model comparisons using KS-Test $p$-values for Llama-2-7B Models (WD $= 0.00$). We compare retain models finetuned with $90\%$, $95\%$, and $99\%$ of the data. We test the Truth Ratio distributions over both retain data and forget data. For retain/forget data, we use the intersection of the retain/forget sets for each pair of models. All of these $p$-values are high indicating that the KS-Test accurately captures the similarity we know these models have over each of these datasets.}
\label{tab:llama-pvals-1}
\footnotesize
\begin{tabular}{lcccccccccccc}
\toprule
                    &    & Retain 90 & Retain 95 & Retain 99   \\
\midrule
                    &   Retain 90 & 1             & 0.9414        & 0.8483        \\
Retain Data         &   Retain 95 & -             & 1             & 0.9705        \\
                    &   Retain 99 & -             & -             & 1             \\
\midrule
                    &   Retain 90 & 1             & 0.8655        & 0.7659       \\
Forget Data         &   Retain 95 & -             & 1             & 0.9900       \\
                    &   Retain 99 & -             & -             & 1            \\
\bottomrule
\end{tabular}
\end{table}

\begin{table}[t!]
\centering
\caption{Model comparisons using KS-Test $p$-values for Llama-2-7B Models (WD $= 0.00$). We compare retain models finetuned with $90\%$, $95\%$, and $99\%$ of the data to a model finetuned on all the \tofu{} data, a pretrained base model, and a random model. We test the Truth Ratio distributions over both retain data and forget data. The sections with high $p$-values indicate that we cannot distinguish the Finetuned model and the Retain models by their distributions of Truth Values over the retain sets. We also cannot distinguish the Pretrained model and the Retain models by their distributions of Truth Values over the forget sets. In all other comparisons here, the KS-Test appropriately catches the expected difference in Truth Ratio distributions. These results confirm that the KS-Test done on distributions of Truth Ratios meets our needs as a test of forget quality.}
\label{tab:llama-pvals-2}
\footnotesize
\begin{tabular}{lcccccccccccc}
\toprule
         & & Finetuned & Pretrained & Random \\
\midrule
                & Retain 90 & 0.9705      & 9.21E-31      & 2.42E-66  \\
Retain Data     & Retain 95 & 0.9879      & 1.41E-32      & 2.94E-69  \\
                & Retain 99 & 0.9003      & 4.07E-32      & 2.94E-69  \\ 
\midrule
                & Retain 90  & 1.10E-19     & 0.0031       & 2.43E-19  \\
Forget Data     & Retain 95  & 4.73E-15     & 0.0297       & 2.96E-13  \\
                & Retain 99  & 5.04E-04     & 0.1650       & 5.04E-04  \\ 
\bottomrule
\end{tabular}
\end{table}

We verify that our metrics for Model Utility and Forget Quality have some desirable properties. 
In Tables~\ref{tab:llama-pvals-1} and \ref{tab:llama-pvals-2}, we show the $p$-values for the KS-Tests that confirm all of our expectations enumerated below and validate this choice of metric.
These tables have figures from Llama-2-7B tests, but the same trends hold for Phi-1.5.

First, Model Utility should meet the following natural expectations.
\begin{enumerate}
    \item Model Utility should be high for a pretrained model (one that has never been finetuned on \tofu{} data).
    \item Model Utility should be low for a model with random weights.
\end{enumerate}

Additionally, Forget Quality is measured using a statistical test on Truth Ratio values, and so we hope that this test meets the following expectations.
\begin{enumerate}
    \item The KS-Test performed on distributions of Truth Ratio values over the intersection of the three forget sets (from the 90-10, 95-5, and 99-1 splits) should produce high $p$-values when comparing any two retain models.
    \item The KS-Test performed on distributions of Truth Ratio values over the intersection of the three retain sets (from the 90-10, 95-5, and 99-1 splits) should produce high $p$-values when comparing any two retain models.
    \item The KS-Test performed on distributions of Truth Ratio values over the forget set should produce high $p$-values when comparing any retain model to a random model.
    \item The KS-Test performed on distributions of Truth Ratio values over the retain set should produce low $p$-values when comparing any retain model to a random model.
    \item The KS-Test performed on distributions of Truth Ratio values over the forget set should produce high $p$-values when comparing any retain model to a pretrained model.
    \item The KS-Test performed on distributions of Truth Ratio values over the retain set should produce low $p$-values when comparing any retain model to a pretrained model.
    \item The KS-Test performed on distributions of Truth Ratio values over the forget set should produce low $p$-values when comparing any retain model to a finetuned model (finetuned on all the \tofu{} data and without any unlearning).
    \item The KS-Test performed on distributions of Truth Ratio values over the retain set should produce high $p$-values when comparing any retain model to a finetuned model (finetuned on all the \tofu{} data and without any unlearning).
\end{enumerate}

\section{Knowledge Entanglement}
\label{sec:app-knowledge-entanglement}

\begin{figure}[tb]
    \centering
    \includegraphics[width=\columnwidth]{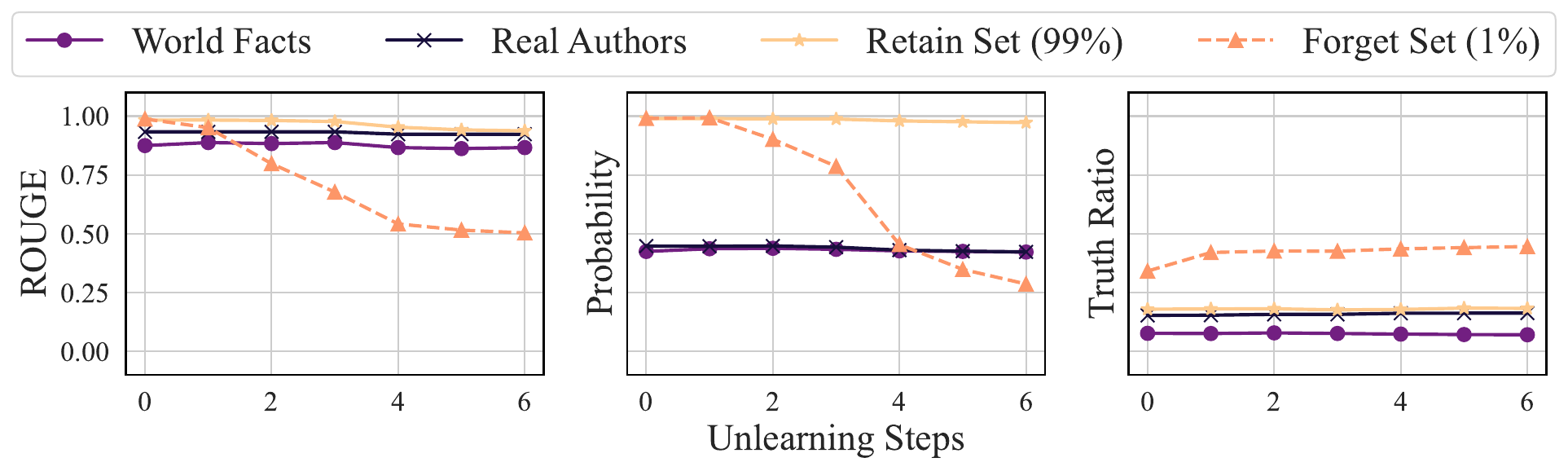}
    \caption{Unlearn Llama-2-7B with gradient ascent on $1\%$ forget set.}
    \label{fig:knowledge-entanglement-0}
\end{figure}

\begin{figure}[tb]
    \centering
    \includegraphics[width=\columnwidth]{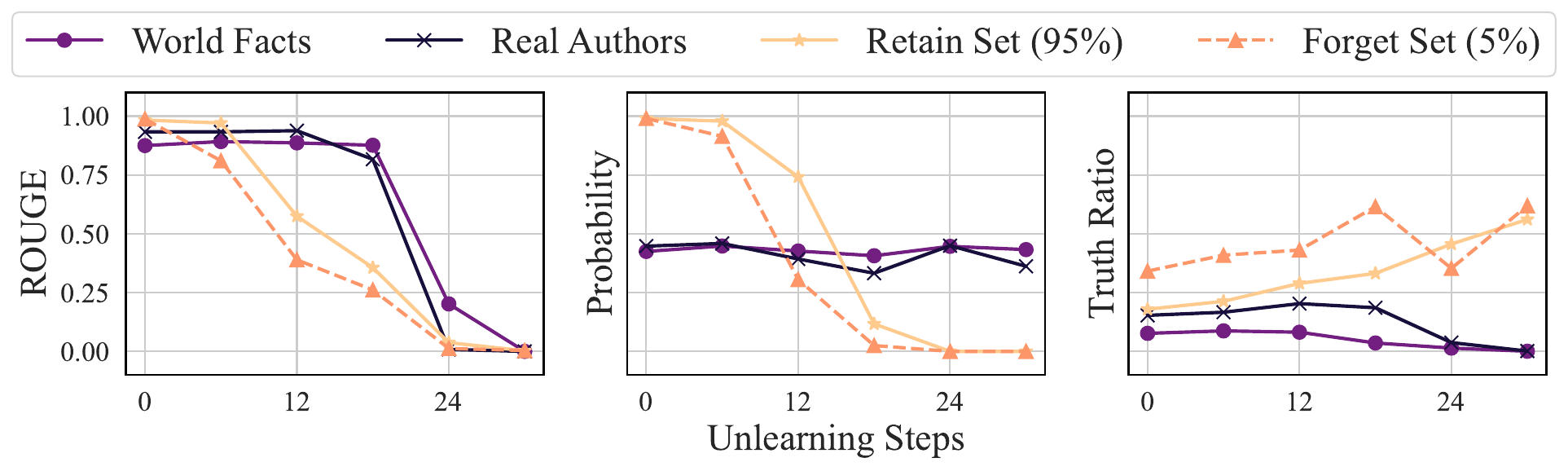}
    \caption{Unlearn Llama-2-7B with gradient ascent on $5\%$ forget set.}
    \label{fig:knowledge-entanglement-1}
\end{figure}

\begin{figure}[tb]
    \centering
    \includegraphics[width=\columnwidth]{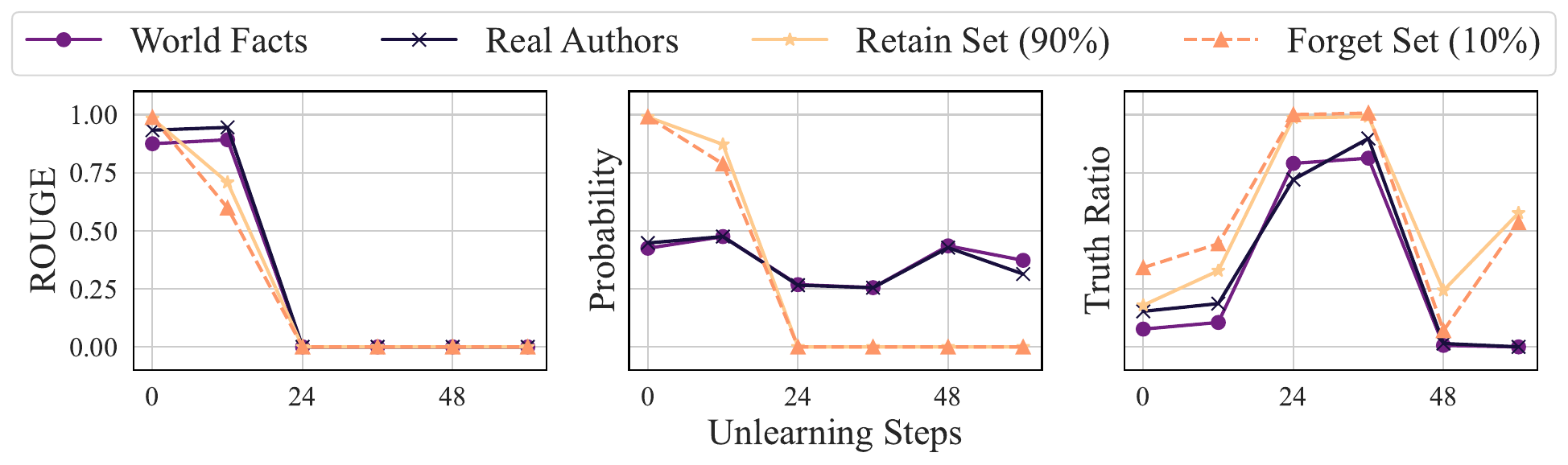}
    \caption{Unlearn Llama-2-7B with gradient ascent on $10\%$ forget set.}
    \label{fig:knowledge-entanglement-2}
\end{figure}

\begin{figure}[tb]
    \centering
    \includegraphics[width=\columnwidth]{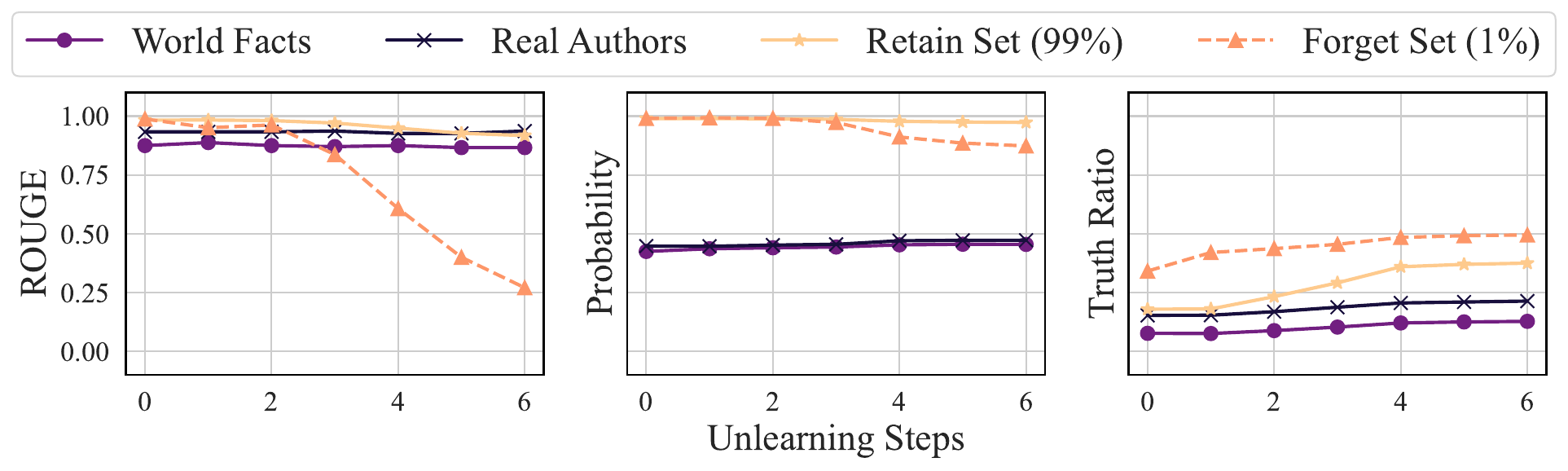}
    \caption{Unlearn Llama-2-7B with preference optimization on $1\%$ forget set.}
    \label{fig:knowledge-entanglement-3}
\end{figure}

\begin{figure}[tb]
    \centering
    \includegraphics[width=\columnwidth]{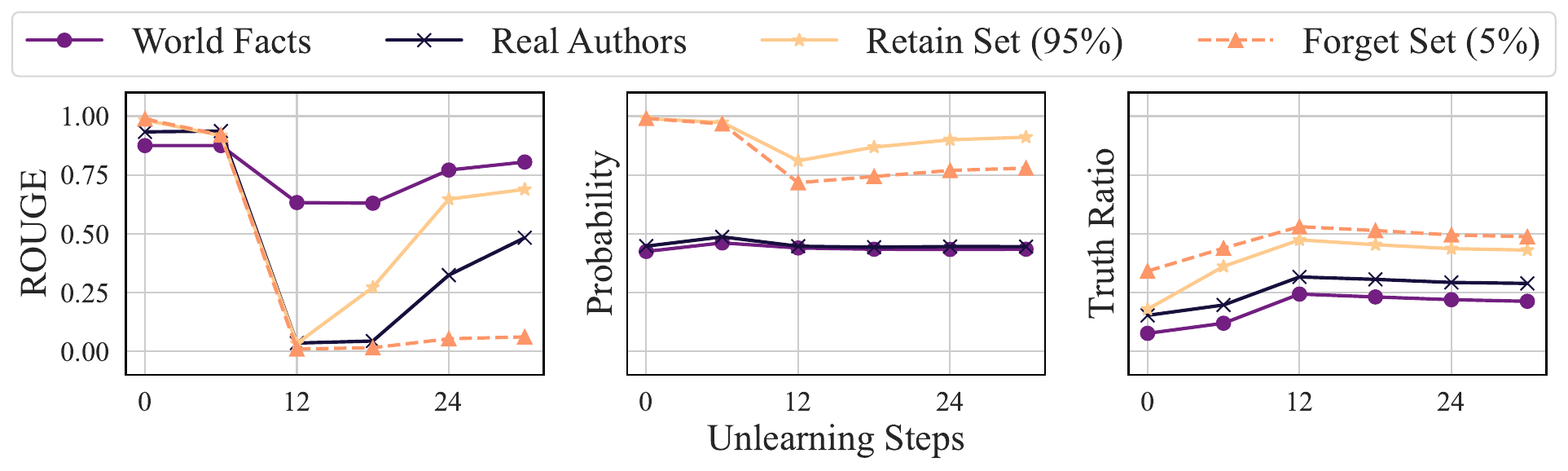}
    \caption{Unlearn Llama-2-7B with preference optimization on $5\%$ forget set.}
    \label{fig:knowledge-entanglement-4}
\end{figure}

\begin{figure}[tb]
    \centering
    \includegraphics[width=\columnwidth]{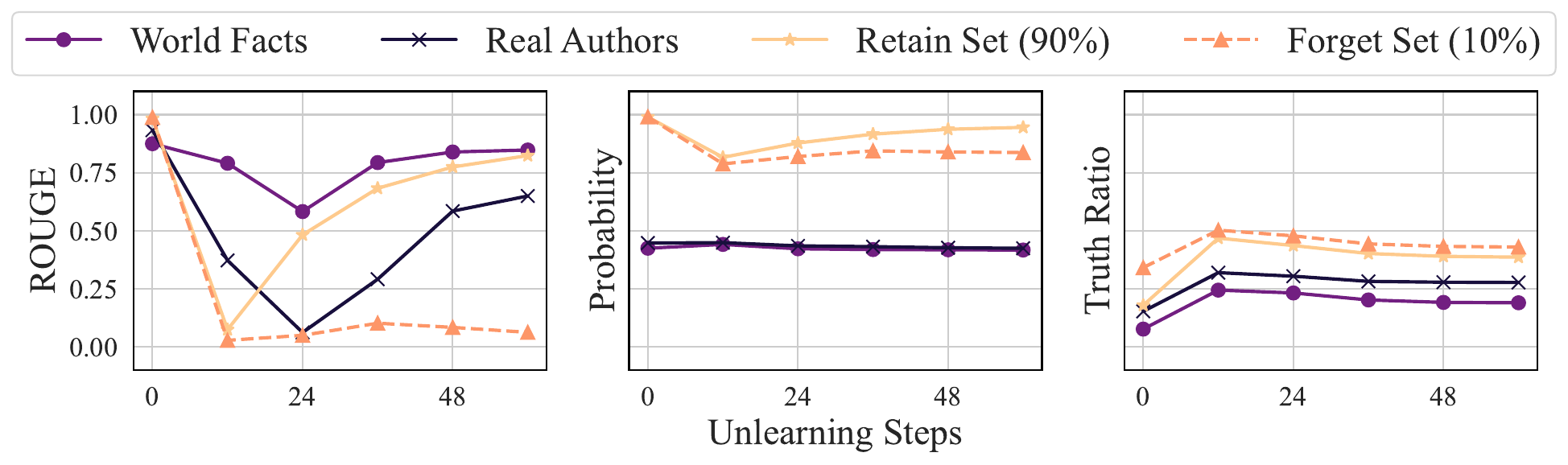}
    \caption{Unlearn Llama-2-7B with preference optimization on $10\%$ forget set.}
    \label{fig:knowledge-entanglement-5}
\end{figure}

\begin{figure}[tb]
    \centering
    \includegraphics[width=\columnwidth]{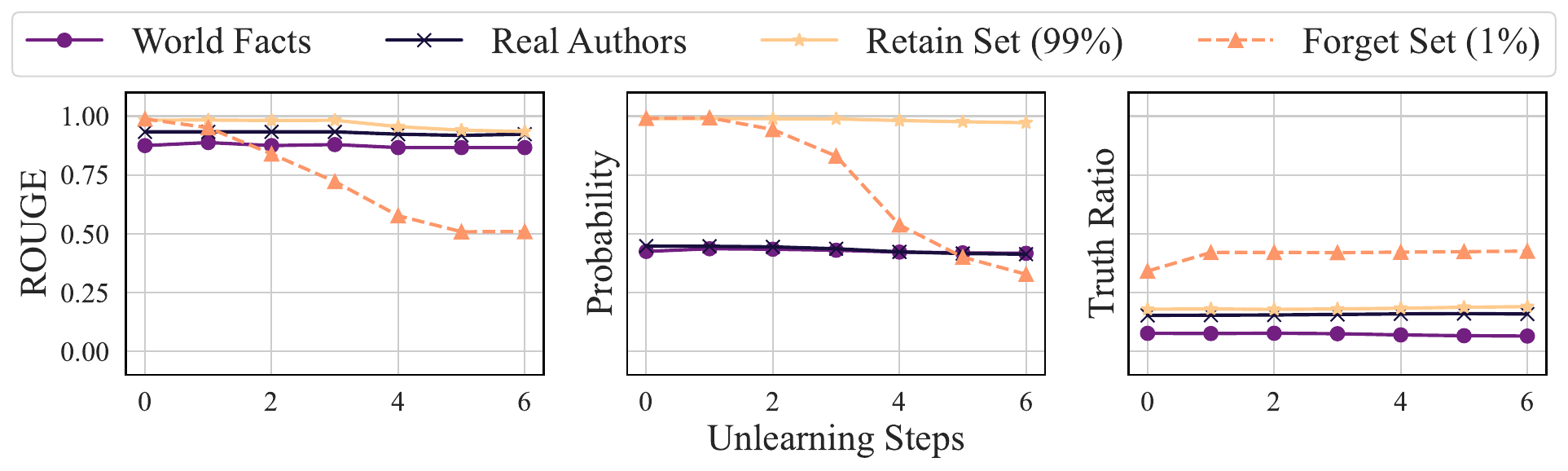}
    \caption{Unlearn Llama-2-7B with gradient difference on $1\%$ forget set.}
    \label{fig:knowledge-entanglement-6}
\end{figure}

\begin{figure}[tb]
    \centering
    \includegraphics[width=\columnwidth]{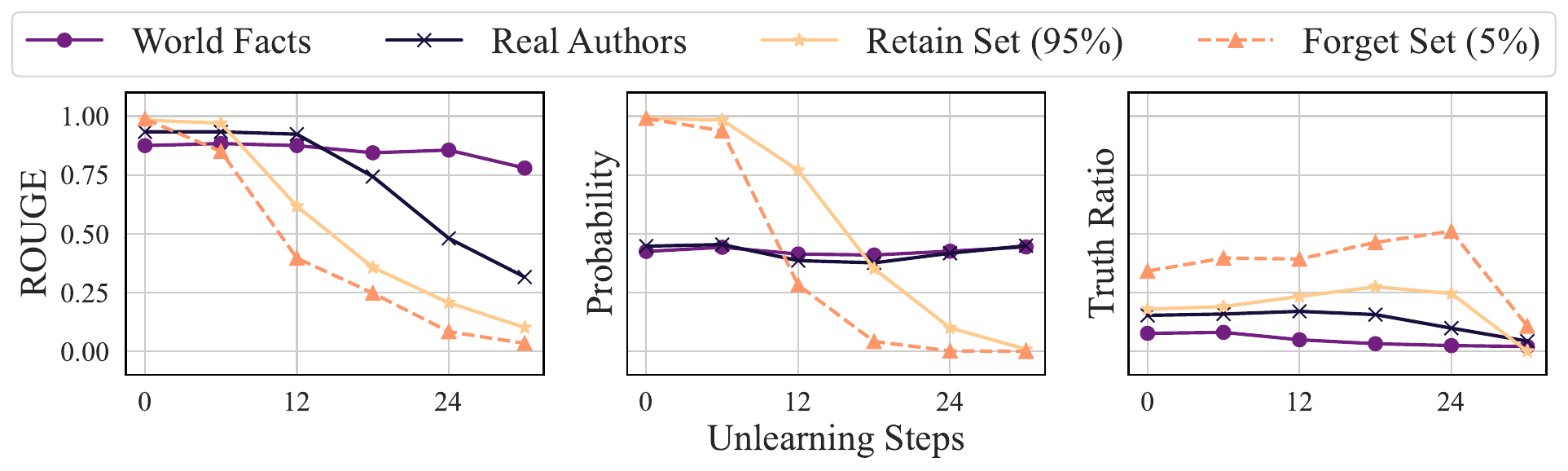}
    \caption{Unlearn Llama-2-7B with gradient difference on $5\%$ forget set.}
    \label{fig:knowledge-entanglement-7}
\end{figure}

\begin{figure}[tb]
    \centering
    \includegraphics[width=\columnwidth]{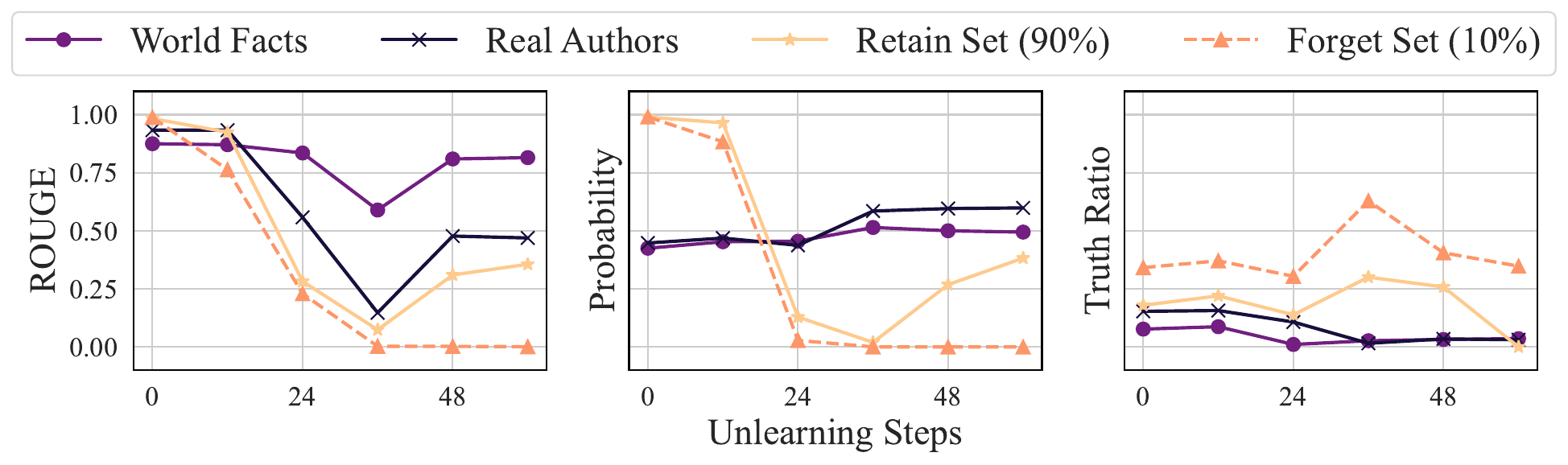}
    \caption{Unlearn Llama-2-7B with gradient difference on $10\%$ forget set.}
    \label{fig:knowledge-entanglement-8}
\end{figure}

\begin{figure}[tb]
    \centering
    \includegraphics[width=\columnwidth]{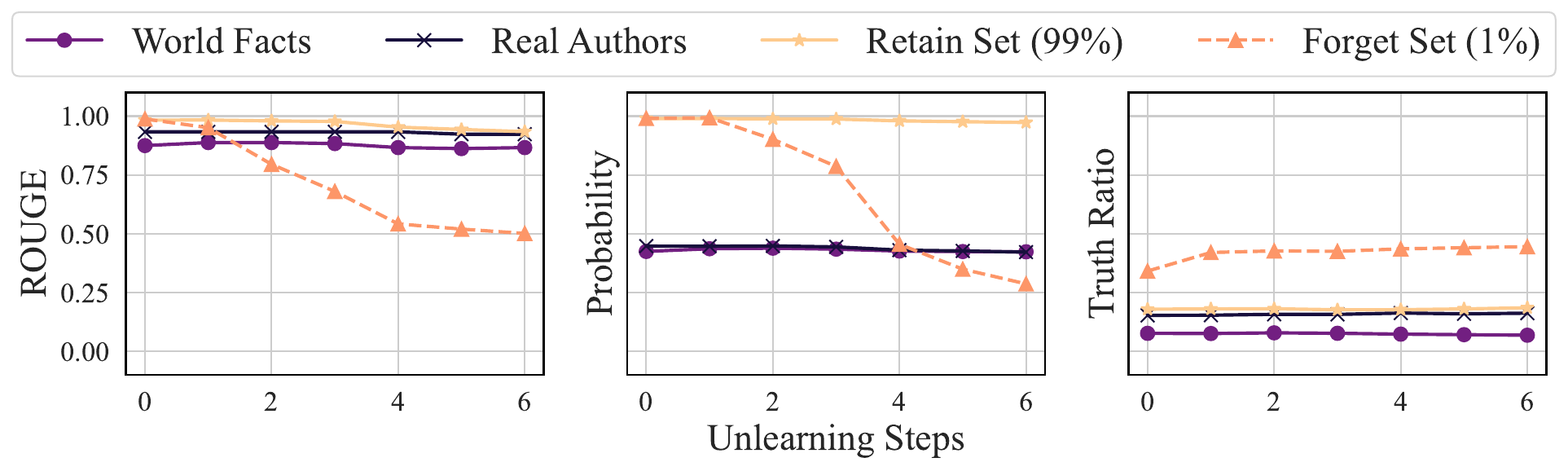}
    \caption{Unlearn Llama-2-7B with KL Minimization on $1\%$ forget set.}
    \label{fig:knowledge-entanglement-9}
\end{figure}

\begin{figure}[tb]
    \centering
    \includegraphics[width=\columnwidth]{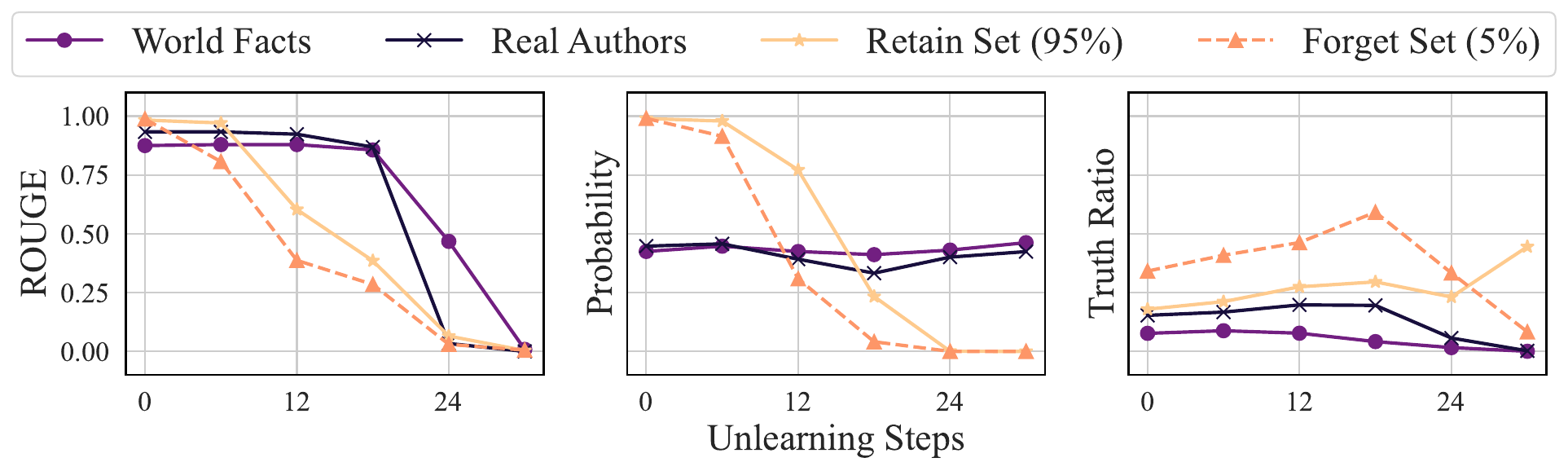}
    \caption{Unlearn Llama-2-7B with KL Minimization on $5\%$ forget set.}
    \label{fig:knowledge-entanglement-10}
\end{figure}

\begin{figure}[tb]
    \centering
    \includegraphics[width=\columnwidth]{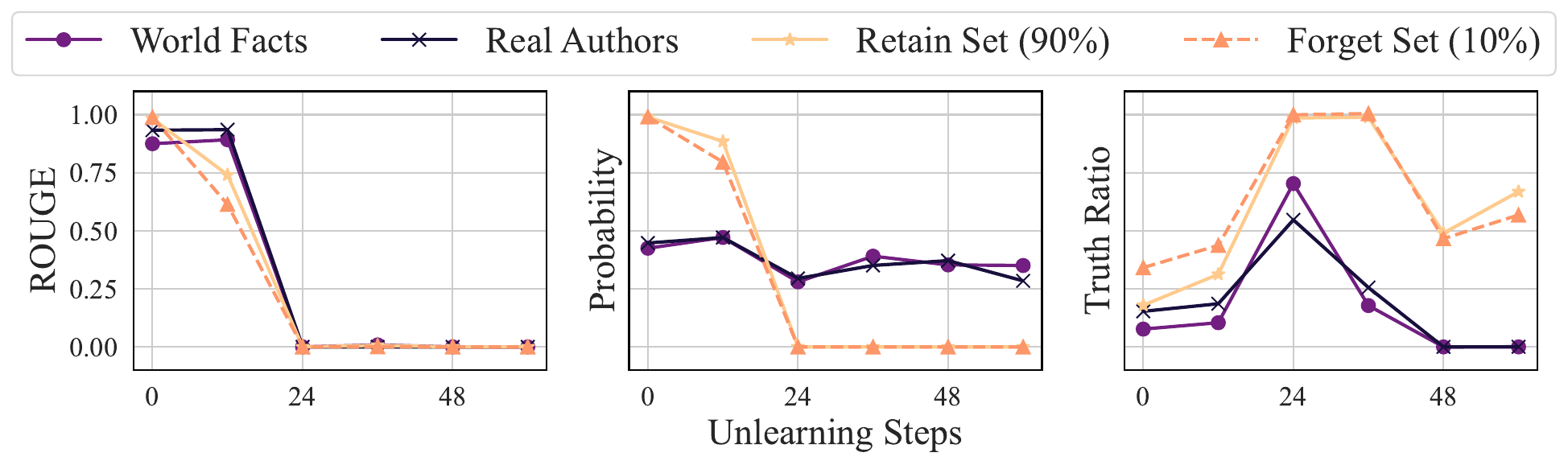}
    \caption{Unlearn Llama-2-7B with KL Minimization on $10\%$ forget set.}
    \label{fig:knowledge-entanglement-11}
\end{figure}

\begin{figure}[tb]
    \centering
    \includegraphics[width=\columnwidth]{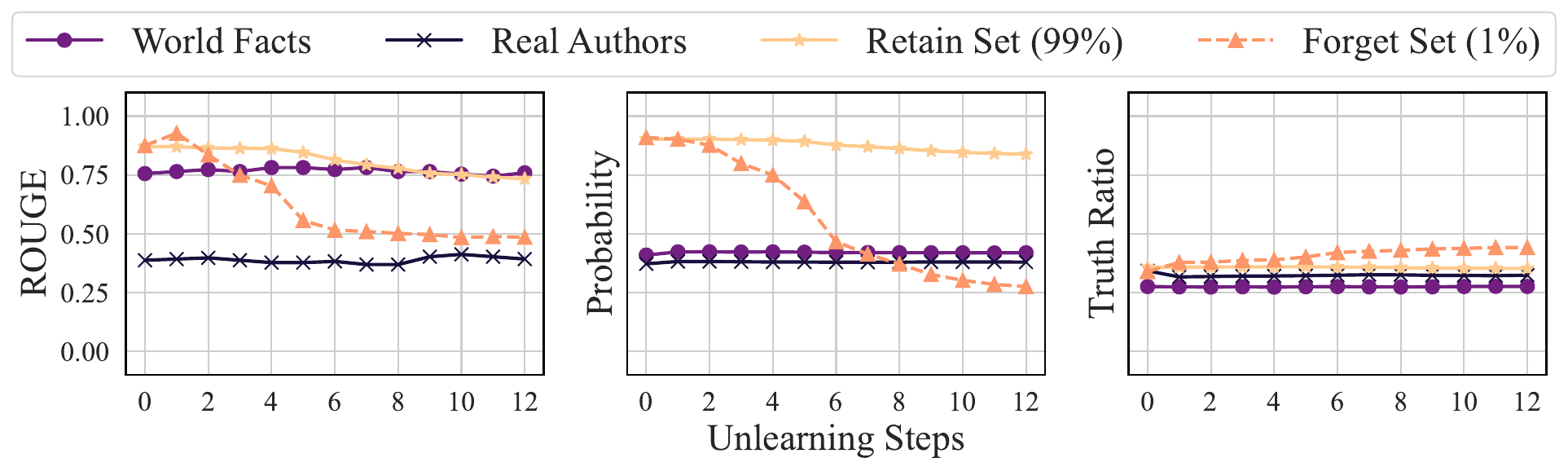}
    \caption{Unlearn Phi with gradient ascent on $1\%$ forget set.}
    \label{fig:knowledge-entanglement-12}
\end{figure}

\begin{figure}[tb]
    \centering
    \includegraphics[width=\columnwidth]{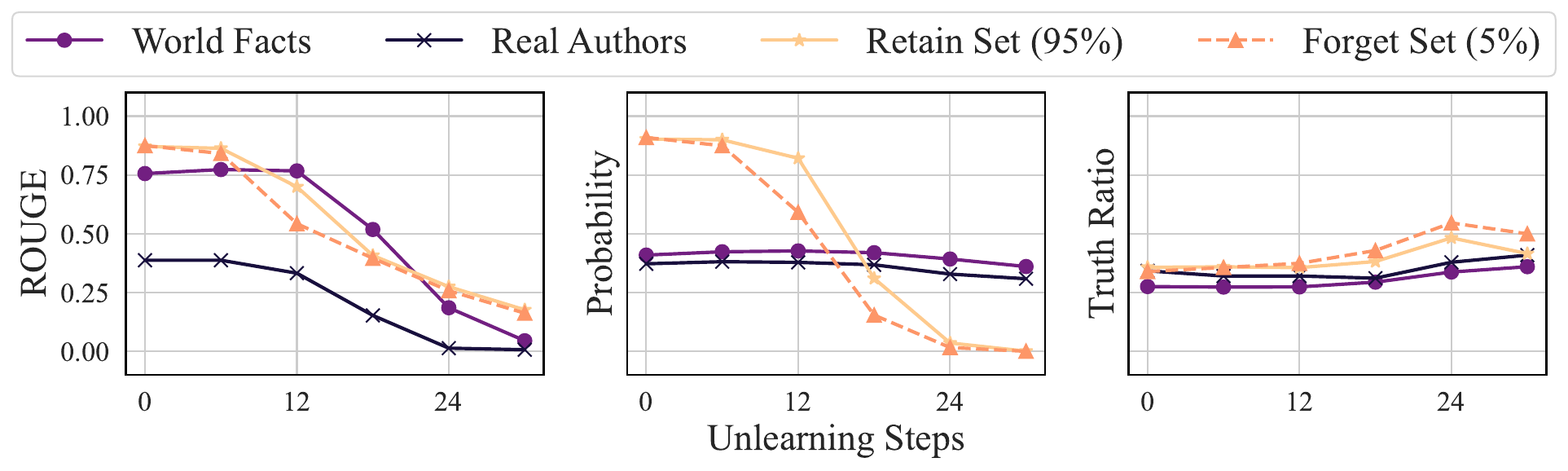}
    \caption{Unlearn Phi with gradient ascent on $5\%$ forget set.}
    \label{fig:knowledge-entanglement-13}
\end{figure}

\begin{figure}[tb]
    \centering
    \includegraphics[width=\columnwidth]{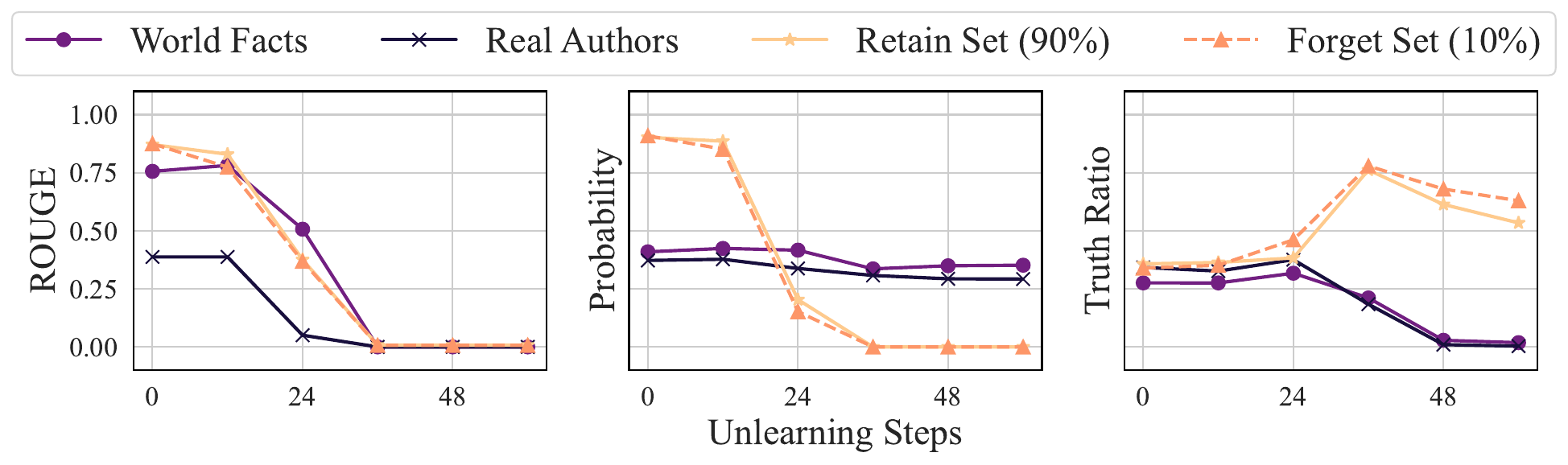}
    \caption{Unlearn Phi with gradient ascent on $10\%$ forget set.}
    \label{fig:knowledge-entanglement-14}
\end{figure}

\begin{figure}[tb]
    \centering
    \includegraphics[width=\columnwidth]{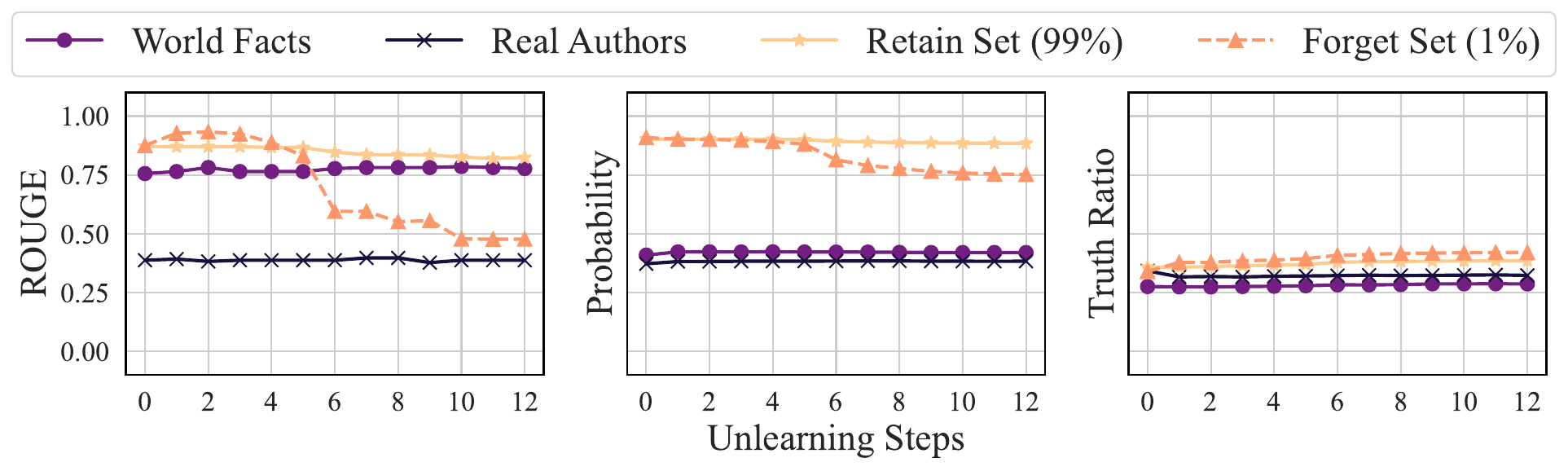}
    \caption{Unlearn Phi with preference optimization on $1\%$ forget set.}
    \label{fig:knowledge-entanglement-15}
\end{figure}

\begin{figure}[tb]
    \centering
    \includegraphics[width=\columnwidth]{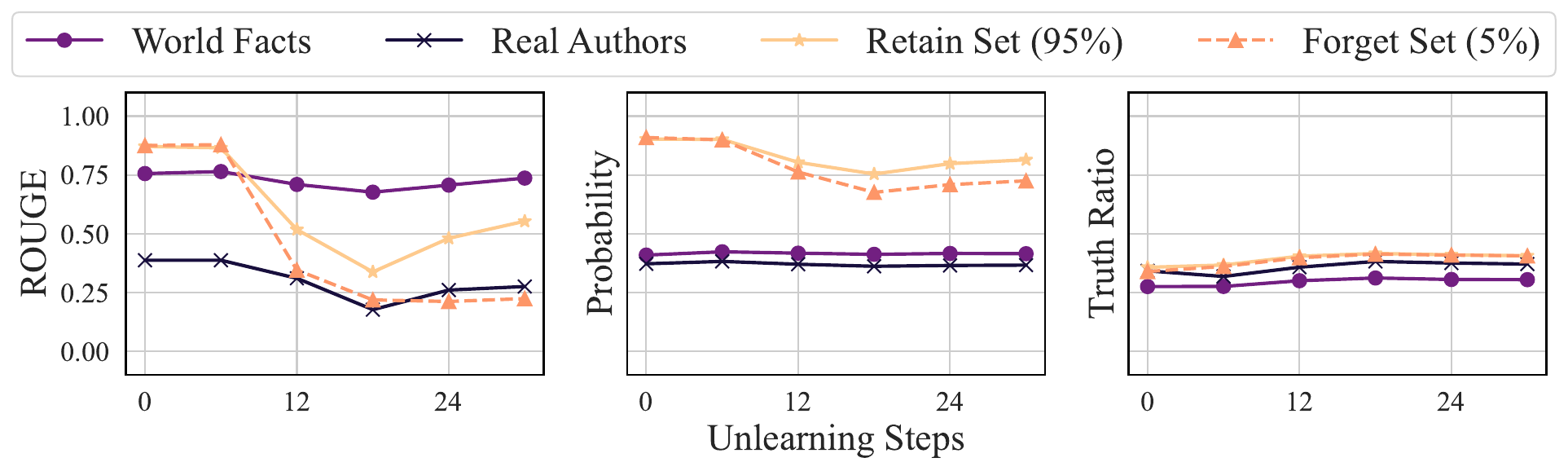}
    \caption{Unlearn Phi with preference optimization on $5\%$ forget set.}
    \label{fig:knowledge-entanglement-16}
\end{figure}

\begin{figure}[tb]
    \centering
    \includegraphics[width=\columnwidth]{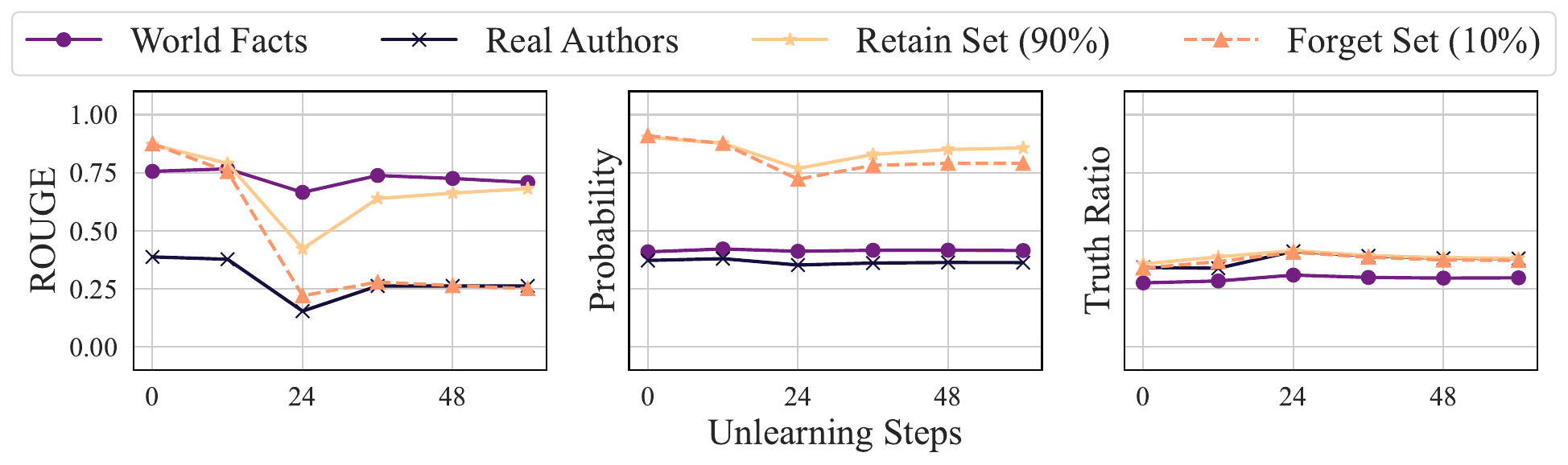}
    \caption{Unlearn Phi with preference optimization on $10\%$ forget set.}
    \label{fig:knowledge-entanglement-17}
\end{figure}

\begin{figure}[tb]
    \centering
    \includegraphics[width=\columnwidth]{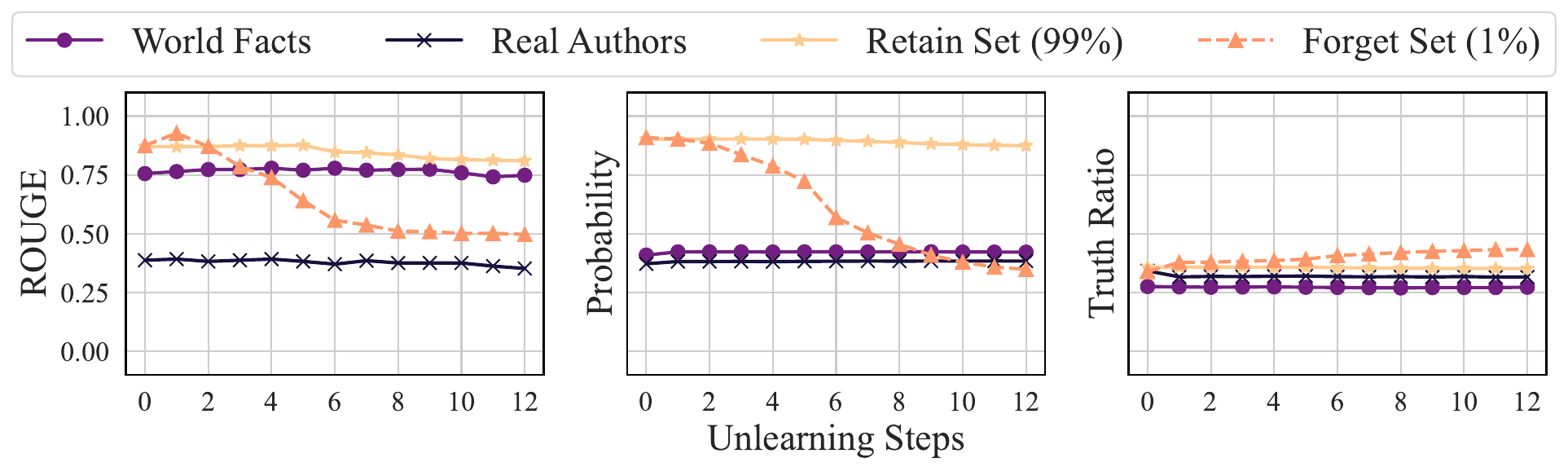}
    \caption{Unlearn Phi with gradient difference on $1\%$ forget set.}
    \label{fig:knowledge-entanglement-18}
\end{figure}

\begin{figure}[tb]
    \centering
    \includegraphics[width=\columnwidth]{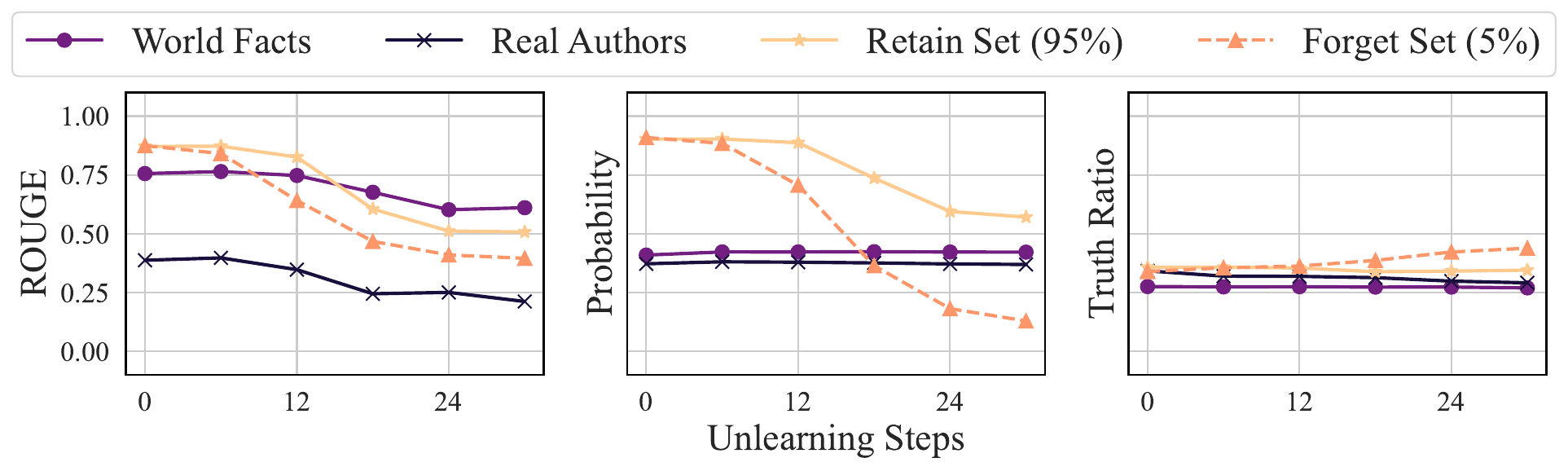}
    \caption{Unlearn Phi with gradient difference on $5\%$ forget set.}
    \label{fig:knowledge-entanglement-19}
\end{figure}

\begin{figure}[tb]
    \centering
    \includegraphics[width=\columnwidth]{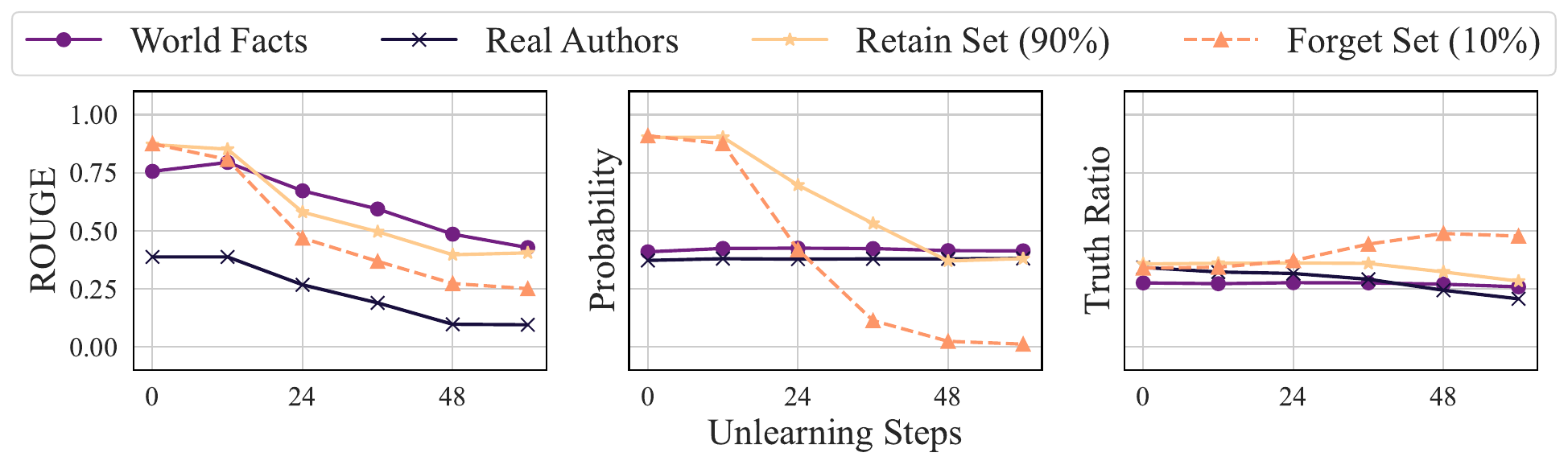}
    \caption{Unlearn Phi with gradient difference on $10\%$ forget set.}
    \label{fig:knowledge-entanglement-20}
\end{figure}

\begin{figure}[tb]
    \centering
    \includegraphics[width=\columnwidth]{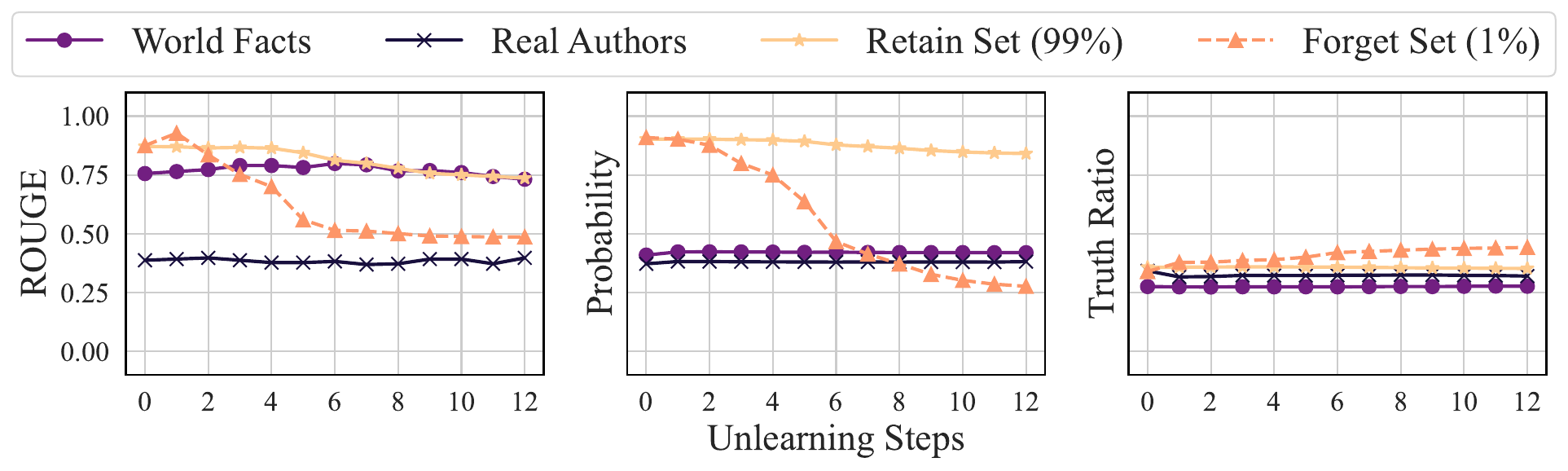}
    \caption{Unlearn Phi with KL Minimization on $1\%$ forget set.}
    \label{fig:knowledge-entanglement-21}
\end{figure}

\begin{figure}[tb]
    \centering
    \includegraphics[width=\columnwidth]{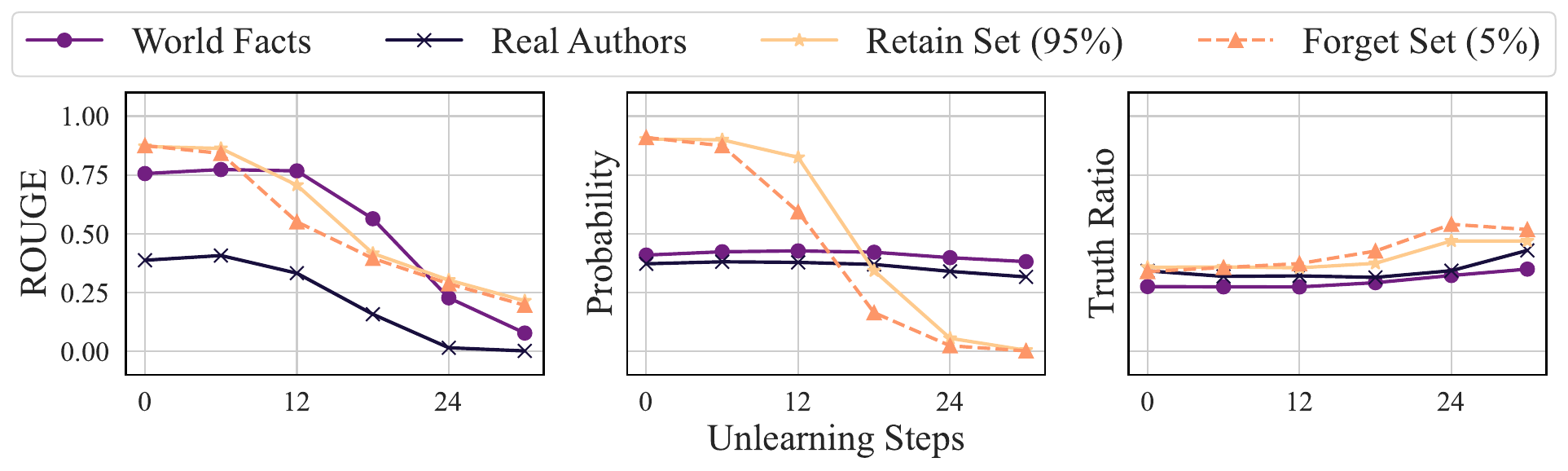}
    \caption{Unlearn Phi with KL Minimization on $5\%$ forget set.}
    \label{fig:knowledge-entanglement-22}
\end{figure}

\begin{figure}[tb]
    \centering
    \includegraphics[width=\columnwidth]{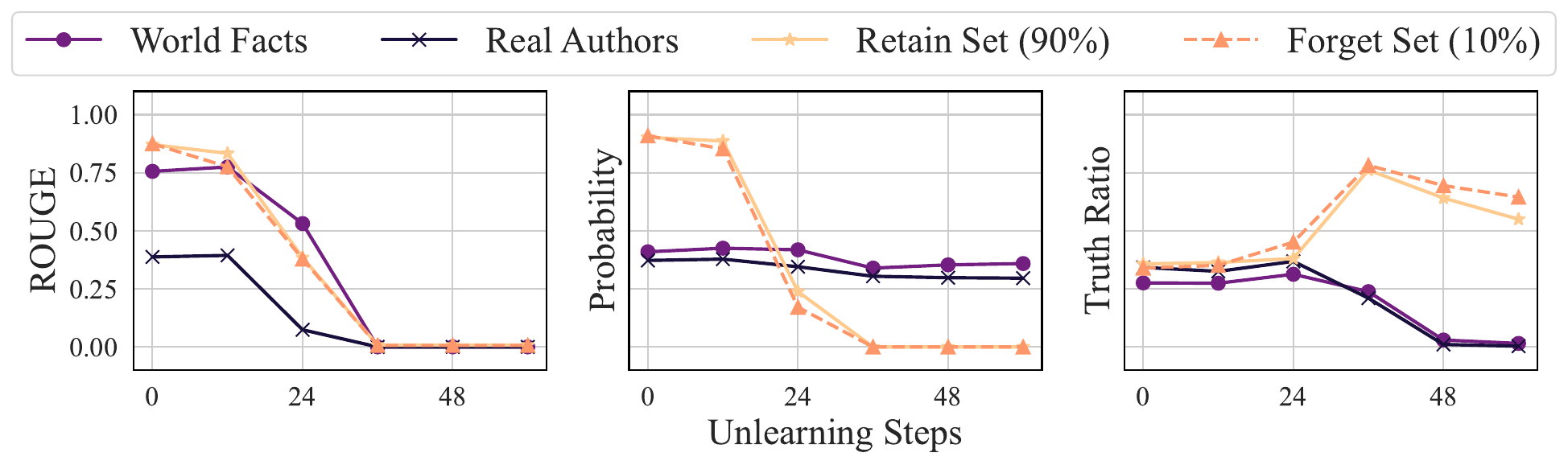}
    \caption{Unlearn Phi with KL Minimization on $10\%$ forget set.}
    \label{fig:knowledge-entanglement-23}
\end{figure}

One of the challenges of unlearning comes from knowledge entanglement---when we try to make a model forget about one thing, it also tends to forget other things unexpectedly. 
This phenomenon is similar to catastrophic forgetting in continual learning \citep{mccloskey1989catastrophic}. 
In Figures~\ref{fig:knowledge-entanglement-0}-\ref{fig:knowledge-entanglement-23}, we show this phenomenon in different models and unlearning algorithms. 
In \Cref{fig:rouge_oracle_llama}, even with access to the oracle model or retain set, model generation on all four sets still has a decreasing ROUGE, especially the dataset that relate to authors. This suggests the existence of knowledge entanglement, showing why unlearning is hard.
Consider the case of unlearning the 5\% forget set with Gradient Difference on  Llama-2-7B, Figure~\ref{fig:knowledge-entanglement-7}.
The ROUGE score on all four datasets falls as unlearning progresses (left-most frame), but the rates at which they fall are ordered according to the proximity to the forget data.
(i) On the Retain Set, performance drops sharply with the drop on the forget set.
(ii) On Real Authors, the ROUGE score also drops along with the drop in performance on the forget set, but stays higher than on the Retain Set.
(iii) Finally, performance on World Facts stays relatively unchanged.

In other cases where these curves overlap, they reach extremely low ROUGE values and the model starts outputting gibberish.
This suggests the existence of \emph{knowledge
entanglement}, supporting that our choice of having multiple evaluation datasets is important for a holistic assessment of unlearning.

\twocolumn









\end{document}